\documentclass[accepted]{uai2026}

\usepackage[american]{babel}

\usepackage{natbib}
\usepackage{mathtools}
\usepackage{booktabs}
\usepackage{tikz}
\usepackage{amsfonts}
\usepackage{graphicx}
\usepackage{algorithm}
\usepackage{algorithmic}
\usepackage{subcaption}

\usepackage{amsopn}
\usepackage{placeins}
\usepackage{stfloats}
\DeclareMathOperator{\diag}{diag}
\DeclareMathOperator{\tr}{tr}

\definecolor{linkblue}{rgb}{0.0,0.0,0.55}
\hypersetup{
  colorlinks=true,
  linkcolor=black,
  citecolor=black,
  urlcolor=linkblue
}

\title{Conditional neural control variates for variance reduction in
  Bayesian inverse problems}

\author[1]{\href{mailto:<alisk@ucf.edu>?Subject=Your UAI 2026 paper}{Ali~Siahkoohi}{}}
\author[2]{Hyunwoo~Oh}
\affil[1]{
    Department of Computer Science\\
    University of Central Florida\\
    Orlando, Florida, USA
}
\affil[2]{
    Department of Physics and Maryland Center for Fundamental Physics\\
    University of Maryland\\
    College Park, Maryland, USA
}

\begin{document}
\maketitle

\begin{abstract}
  Bayesian inference for inverse problems involves computing
  expectations under posterior distributions---e.g., posterior means,
  variances, or predictive quantities---typically via Monte Carlo~(MC)
  estimation. When the quantity of interest varies significantly under the posterior,
  accurate estimates demand many samples---a cost often
  prohibitive for partial differential equation-constrained problems.
  To address this challenge, we introduce conditional neural control
  variates, a modular method that learns amortized control
  variates from joint model--data samples to reduce the variance of MC
  estimators. To scale to high-dimensional problems, we leverage
  Stein's identity to design an architecture based on an ensemble of
  hierarchical coupling layers with tractable Jacobian trace
  computation. Training requires:
  (i)~samples from the joint distribution of unknown parameters and
  observed data; and (ii)~the posterior score function, which can be
  computed from physics-based likelihood evaluations, neural operator
  surrogates, or learned generative models such as conditional
  normalizing flows.
  Once trained, the control variates generalize across observations
  without retraining.
  We validate our approach on stylized and
  partial differential equation-constrained Darcy flow inverse problems,
  outperforming classical Stein control variates and achieving
  substantial variance reduction, even when the analytical score is
  replaced by a learned surrogate.
\end{abstract}

\section{Introduction}\label{sec:intro}

Quantifying uncertainty in the solution of inverse problems is of
fundamental importance in many areas of science and
engineering~\citep{Tarantola2005, Stuart2010}. Due to noise and
ill-posedness, a single point estimate fails to capture the
non-uniqueness of the solution. The Bayesian framework addresses this
by characterizing the full posterior distribution over solutions
consistent with the observed data and prior knowledge. Posterior
samples then provide expectations of quantities of interest---e.g.,
posterior means, variances, and credible intervals.
Computing these expectations requires evaluating integrals of the form
$\mathbb{E}_{p_{\text{post}}(\mathbf{x}|\mathbf{y})}[h(\mathbf{x})]$,
which are typically intractable in closed form. While Monte Carlo~(MC)
estimation replaces the integral with a sample
average~\citep{RobertCasella2004}, its
variance---$\text{Var}(h(\mathbf{x}))/N$ for $N$ i.i.d.\ samples---can be prohibitively
large when $h(\mathbf{x})$ varies significantly under the posterior.
In partial differential equation~(PDE)-constrained inverse problems, where each posterior sample
requires a computationally expensive forward solve, this high variance
translates into high computational cost.

Control variates offer a classical remedy~\citep{FiellerHartley1954,
  RobertCasella2004}: given an auxiliary function $g(\mathbf{x})$ with
zero expectation under the target distribution and positive
correlation with $h(\mathbf{x})$, the controlled estimator
$\mathbb{E}_{p_{\text{post}}(\mathbf{x}|\mathbf{y})}[h(\mathbf{x}) - g(\mathbf{x})]$ has lower variance while remaining
unbiased. Stein's identity~\citep{Stein1972} provides a systematic
way to construct such zero-mean functions from the score
function---i.e., the gradient of the log-density---of the target
distribution. Building on this principle, control
functionals~\citep{OatesGirolamiChopin2017} and the zero-variance
principle~\citep{AssarafCaffarel1999} have demonstrated the
effectiveness of score-based control variates for MC integration, and
neural networks have been used to parameterize them in gradient
estimation~\citep{Grathwohl2018}, rendering~\citep{Muller2020},
general MC integration~\citep{Wan2020}, and lattice field
theory~\citep{BedaqueOh2024, Oh2025}. While these methods achieve
impressive variance reduction, they all require retraining for each
new observation $\mathbf{y}$, precluding amortization---a crucial
limitation when retraining for every new observation is computationally
prohibitive.

To address this limitation, we introduce conditional neural control
variates~(CNCV), a modular variance reduction method for MC
estimation of posterior expectations in Bayesian inverse problems. Our
method requires:
(i)~joint samples $(\mathbf{x}, \mathbf{y})$ drawn from the prior
predictive distribution---by sampling $\mathbf{x}$ from the prior and
computing
$\mathbf{y} = \mathcal{F}(\mathbf{x}) +
\boldsymbol{\epsilon}$---samples that are already available in
simulation-based inference
pipelines~\citep{Cranmer2020}; and
(ii)~the posterior score function
$\nabla_\mathbf{x} \log p(\mathbf{x}|\mathbf{y})$, which can come
from physics-based evaluations, neural operator
surrogates~\citep{LiKovachki2021}, or learned generative
models~\citep{RezendeMohamed2015, SongErmon2019, HoJainAbbeel2020, Papamakarios2021,
  BaldassariEtAl2023}. Once
trained, the control variates apply to any new
observation $\mathbf{y}$ without retraining, integrating seamlessly
with simulation-based inference techniques. In practice, CNCV drops
into such a pipeline in three steps: (1)~train an amortized
posterior---for instance a conditional normalizing flow---on the joint
samples; (2)~train CNCV on the same samples using that posterior's
score; and (3)~at inference, draw samples, evaluate the score, and
subtract the learned control variate. While we focus on Bayesian
inverse problems, the method applies wherever Monte Carlo estimation is
the bottleneck and a smooth conditional score is
available---for example, reinforcement-learning policy
gradients~\citep{Greensmith2004}, latent uncertainty quantification
with functional autoencoders~\citep{Bunker2025}, and amortized
uncertainty quantification for digital
twins~\citep{Herrmann2023digitaltwin}.

A further challenge arises when scaling Stein-based control variates
to high dimensions: computing the divergence of a general neural
network---required by the construction in
Section~\ref{sec:cncv}---involves $\mathcal{O}(d)$ backward passes
or stochastic trace estimation~\citep{Hutchinson1990}. To address
this, we introduce an ensemble of hierarchical affine coupling layers
inspired by HINT~\citep{kruse2021hint}, each operating on a random
permutation of the input dimensions. The triangular Jacobian structure
enables exact divergence computation in a single forward pass. Each
ensemble member captures different cross-dimension interactions
through its permutation,
and averaging yields control variates with high correlation across all
components of the quantity of interest.

Our main contributions are:
(i)~we extend neural control variates to the conditional setting,
yielding amortized control variates that integrate with any posterior
sampler and score source;
(ii)~we introduce an ensemble of hierarchical coupling
layers with random input permutations that
enable exact divergence computation; and
(iii)~we validate CNCV on four inverse problems of increasing
complexity---Gaussian, Rosenbrock, nonlinear, and PDE-constrained
Darcy flow---where it outperforms classical Stein control variates and
reduces variance across observations, even under learned scores and
non-Gaussian noise.

\section{Bayesian inverse problems}\label{sec:background}

We are concerned with estimating an unknown parameter
$\mathbf{x} \in \mathcal{X} \subset \mathbb{R}^d$ from noisy
observations
$\mathbf{y} \in \mathcal{Y} \subset \mathbb{R}^m$ related through the
forward model
$\mathbf{y} = \mathcal{F}(\mathbf{x}) + \boldsymbol{\epsilon}$,
where
$\mathcal{F}: \mathcal{X} \to \mathcal{Y}$ is the forward operator
and $\boldsymbol{\epsilon} \sim \mathcal{N}(\mathbf{0}, \sigma^2
\mathbf{I})$ is additive Gaussian noise. In the Bayesian
framework~\citep{Tarantola2005, KaipioSomersalo2006}, the posterior
distribution is given by Bayes' rule:
\begin{equation}\label{eq:posterior}
  p_{\text{post}}(\mathbf{x} | \mathbf{y})
  \propto p_{\text{like}}(\mathbf{y}|\mathbf{x}) \,
    p_{\text{prior}}(\mathbf{x}),
\end{equation}
where
$p_{\text{like}}(\mathbf{y}|\mathbf{x})$ is the likelihood,
$p_{\text{prior}}(\mathbf{x})$ encodes prior beliefs on the unknown,
and the normalizing constant is the marginal likelihood $p(\mathbf{y})$.
For brevity, we write
$p(\mathbf{x}|\mathbf{y}) \equiv
p_{\text{post}}(\mathbf{x}|\mathbf{y})$ throughout.
Since the posterior is generally intractable, we must resort to
sampling-based methods to extract information from it.

\subsection{Sampling the posterior}\label{sec:score}

Extracting information from the posterior requires access to
samples---and many modern posterior sampling methods rely on the
\emph{score function}
$\nabla_\mathbf{x} \log p(\mathbf{x}|\mathbf{y})$, either explicitly
or implicitly. By Bayes' rule, the score decomposes as
$\nabla_\mathbf{x} \log p(\mathbf{x}|\mathbf{y})
= \nabla_\mathbf{x} \log p_{\text{like}}(\mathbf{y}|\mathbf{x})
+ \nabla_\mathbf{x} \log p_{\text{prior}}(\mathbf{x})$.
Gradient-based MCMC methods---such as MALA and
HMC~\citep{RobertsRosenthal1998, Neal2011}---use the score to
construct Markov chains that converge to the posterior, but produce
correlated samples and require burn-in. Conditional generative
models offer an alternative: conditional normalizing
flows~(CNFs;~\citealp{RezendeMohamed2015, Papamakarios2021}) learn an invertible
map from a base distribution to the posterior, providing both i.i.d.\
samples and an explicit density from which the score is obtained by
automatic differentiation. Diffusion
models~\citep{SongErmon2019, HoJainAbbeel2020, BaldassariEtAl2023}
directly parameterize the score via denoising score matching. When the
forward operator $\mathcal{F}$ is differentiable---or when its adjoint
is available---the likelihood score can also be
evaluated analytically~\citep{LouboutinEtAl2023}; when $\mathcal{F}$ is expensive, a
differentiable surrogate~\citep{LiKovachki2021} can approximate it.
In all cases, the score of the sampling distribution is available as a
byproduct. Our control variate construction
(Section~\ref{sec:cncv}) exploits this observation, requiring only
that the score corresponds to the sampling distribution.

\subsection{Monte Carlo estimation and control variates}

Once posterior samples are available, the primary computational task is
estimating expectations of quantities of interest
$h\colon \mathcal{X} \to \mathbb{R}$:
\begin{equation}\label{eq:expectation}
  \mathbb{E}_{p_{\text{post}}(\mathbf{x}|\mathbf{y})}[h(\mathbf{x})]
  = \int_{\mathcal{X}} h(\mathbf{x}) \,
  p_{\text{post}}(\mathbf{x}|\mathbf{y}) \, d\mathbf{x}.
\end{equation}
These expectations provide statistical summaries---e.g., the posterior
mean ($h(\mathbf{x}) = \mathbf{x}$), posterior variance
($h(\mathbf{x}) = (\mathbf{x} - \boldsymbol{\mu}_{\text{post}})^2$),
or predictive quantities. Given $N$ samples
$\{\mathbf{x}_i\}_{i=1}^N \sim p(\mathbf{x}|\mathbf{y})$,
the MC estimator
$\widehat{\mathbb{E}}[h(\mathbf{x})]
= \frac{1}{N} \sum_{i=1}^N h(\mathbf{x}_i)$
is unbiased with variance $\text{Var}(h(\mathbf{x}))/N$. When
$\text{Var}(h(\mathbf{x}))$ is large, accurate estimates demand many
samples.
A control variate
$g\colon \mathcal{X} \times \mathcal{Y} \to \mathbb{R}$ satisfying
$\mathbb{E}_{p(\mathbf{x}|\mathbf{y})}[g(\mathbf{x},\mathbf{y})] =
0$ yields the controlled estimator
\begin{equation}\label{eq:cv}
  \widetilde{\mathbb{E}}[h(\mathbf{x})]
  = \frac{1}{N} \sum_{i=1}^N
  \bigl(h(\mathbf{x}_i) - g(\mathbf{x}_i, \mathbf{y})\bigr),
\end{equation}
which is also unbiased, with per-sample variance
$\text{Var}(h - g) = \text{Var}(h) + \text{Var}(g) - 2\,\text{Cov}(h, g)$.
When $g$ is positively correlated with $h$ and has moderate variance,
$\text{Var}(h - g) < \text{Var}(h)$---i.e., the control variate
reduces the per-sample variance. When the samples
$\{\mathbf{x}_i\}_{i=1}^N$ are i.i.d.\ draws from the posterior---e.g.,
from a conditional generative model or independent importance
sampling---the terms
$h(\mathbf{x}_i) - g(\mathbf{x}_i, \mathbf{y})$ are themselves
i.i.d., and the estimator variance is $\text{Var}(h - g)/N$. The
control variate therefore reduces the constant factor from
$\text{Var}(h)$ to $\text{Var}(h - g)$ while preserving the $1/N$
convergence rate. The challenge is constructing such zero-mean
functions $g$ systematically; our approach
(Section~\ref{sec:cncv}) uses the score from the sampling distribution
to build $g$ via Stein's identity.

\section{Conditional neural control variates}\label{sec:cncv}

The key
insight is to combine Stein's identity---which guarantees zero
expectation by construction---with a hierarchical coupling layer
architecture that makes the required divergence computation tractable.

\subsection{Control variate construction via Stein's identity}

Stein's identity~\citep{Stein1972} states that for any smooth
$\boldsymbol{\phi}\colon \mathbb{R}^d \times \mathbb{R}^m \to
\mathbb{R}^d$ satisfying appropriate boundary conditions
(see Appendix~\ref{app:stein}),
\begin{equation}\label{eq:stein}
  \mathbb{E}_{p(\mathbf{x}|\mathbf{y})}\!\left[
    \nabla_\mathbf{x} \cdot \boldsymbol{\phi}(\mathbf{x},\mathbf{y})
    + \boldsymbol{\phi}(\mathbf{x},\mathbf{y}) \cdot
    \nabla_\mathbf{x} \log p(\mathbf{x}|\mathbf{y})
  \right] = 0.
\end{equation}
The above expression immediately provides a family of control variates
parameterized by $\boldsymbol{\phi}$:
\begin{equation}\label{eq:cv_stein}
  g(\mathbf{x},\mathbf{y})
  = \nabla_\mathbf{x} \cdot \boldsymbol{\phi}(\mathbf{x},\mathbf{y})
  + \boldsymbol{\phi}(\mathbf{x},\mathbf{y}) \cdot
  \nabla_\mathbf{x} \log p(\mathbf{x}|\mathbf{y}),
\end{equation}
where
$\nabla_\mathbf{x} \cdot \boldsymbol{\phi} = \sum_{i=1}^d
\partial\phi_i / \partial x_i$ denotes the divergence. By
construction,
$\mathbb{E}_{p(\mathbf{x}|\mathbf{y})}[g(\mathbf{x},\mathbf{y})] =
0$ for all $\mathbf{y}$, ensuring that $g$ could be used as a control variate.

We parameterize $\boldsymbol{\phi}$ by a neural network
$\boldsymbol{\phi}_\theta\colon \mathbb{R}^{d+m} \to \mathbb{R}^d$;
substituting into equation~\eqref{eq:cv_stein} yields $g_\theta$.
Unfortunately, computing
$\nabla_\mathbf{x} \cdot \boldsymbol{\phi}_\theta
= \sum_{i=1}^d \partial\phi_{\theta,i}/\partial x_i$
for a general neural network requires $d$ backward passes or
stochastic trace estimation~\citep{Hutchinson1990}---a cost that
becomes prohibitive in high dimensions
(Appendix~\ref{app:hutchinson} quantifies its accuracy and cost),
motivating the architecture below.

\subsection{Hierarchical coupling layer architecture}

To address the computational challenge of the divergence, we
parameterize
$\boldsymbol{\phi}_\theta$ using the hierarchical invertible neural
transport (HINT) architecture of~\citet{kruse2021hint}, which
organizes affine coupling layers~\citep{dinh2017density} in a
recursive binary tree. At each node, the input is split into two
halves $\mathbf{x} = [\mathbf{x}_1, \mathbf{x}_2]$ with
$\mathbf{x}_1 \in \mathbb{R}^{d_1}$,
$\mathbf{x}_2 \in \mathbb{R}^{d_2}$, and transformed
via an affine coupling:
\begin{equation}\label{eq:coupling}
  \boldsymbol{\phi}_\theta(\mathbf{x},\mathbf{y})
  = \begin{bmatrix}
    \mathbf{x}_1 \\
    \mathbf{s}_\theta(\mathbf{x}_1,\mathbf{y}) \odot \mathbf{x}_2
    + \mathbf{t}_\theta(\mathbf{x}_1,\mathbf{y})
  \end{bmatrix}
\end{equation}
Here, $\mathbf{s}_\theta$ and $\mathbf{t}_\theta$ are neural
networks conditioned on $\mathbf{y}$, and $\odot$ denotes
elementwise multiplication. In the recursive tree, the upper half
is first processed by its subtree, and the resulting output---rather
than the raw $\mathbf{x}_1$---conditions the coupling of the lower
half; the coupled lower half is then processed by its own
subtree. This continues until leaf nodes are reached at a prescribed
depth. The recursive subdivision ensures that every dimension
participates in at least one coupling, producing a lower-triangular
Jacobian with all but one diagonal entry learnable.

At a single coupling node, the triangular structure yields diagonal
elements $\partial\phi_{\theta,i}/\partial x_i = 1$ for $i$ in the
upper partition and
$\partial\phi_{\theta,i}/\partial x_i =
s_{\theta,i}(\mathbf{x}_1,\mathbf{y})$ for $i$ in the lower
partition.
The divergence contribution from one node is thus
\begin{equation}\label{eq:div}
  \sum_{i \in \mathbf{x}_2}
  s_{\theta,i}(\mathbf{x}_1,\mathbf{y}) + d_1,
\end{equation}
where $d_1 = \dim(\mathbf{x}_1)$.
The full-tree Jacobian diagonal is computed recursively:
$\mathbf{d} = [\mathbf{d}_{\text{upper}};\;
\mathbf{s} \odot \mathbf{d}_{\text{lower}}]$, with
$\mathbf{d} = \mathbf{1}$ at leaf nodes (see
Appendix~\ref{app:stein}). The divergence
$\nabla_\mathbf{x} \cdot \boldsymbol{\phi}_\theta = \mathbf{1}^\top
\mathbf{d}$ is exact and obtained in a single forward pass.
This exactness is decisive at scale: at $d{=}100$, matching the
accuracy of the analytical diagonal with a generic black-box
$\boldsymbol{\phi}$ via Hutchinson trace
estimation~\citep{Hutchinson1990} requires roughly $100$ random
probes, still incurs an $8$--$11\%$ relative trace error, and costs
about $100\times$ more per evaluation; the triangular Jacobian
structure is therefore a prerequisite for tractable, low-variance
divergence estimation in high dimensions, not merely a convenience
(Appendix~\ref{app:hutchinson}).

Each hierarchical coupling tree $\boldsymbol{\phi}_\theta$ maps
$\mathbf{x} \in \mathbb{R}^d$ to $\mathbb{R}^d$. Crucially, the
full Jacobian diagonal
$\diag(\nabla_\mathbf{x}\boldsymbol{\phi}_\theta)$ is available from
the recursive forward pass, enabling a \emph{vector-valued} control
variate
for $\mathbf{h}(\mathbf{x}) \in \mathbb{R}^k$ using a single network:
\begin{equation}\label{eq:component_cv}
  \mathbf{g}_\theta(\mathbf{x}, \mathbf{y})
  = \diag\bigl(\nabla_\mathbf{x}\boldsymbol{\phi}_\theta\bigr)
  + \boldsymbol{\phi}_\theta(\mathbf{x}, \mathbf{y})
  \odot \nabla_\mathbf{x} \log p(\mathbf{x}|\mathbf{y}),
\end{equation}
In the above expression, $\odot$ denotes elementwise multiplication
and $k$ denotes the dimension of the quantity of interest. Since the
$j$-th component
$g_{\theta,j} = \partial\phi_{\theta,j}/\partial x_j + \phi_{\theta,j}
\cdot \nabla_{x_j} \log p(\mathbf{x}|\mathbf{y})$ is exactly the
$j$-th term of the product rule expansion
\emph{before} summing over components (see
Appendix~\ref{app:stein}), each component independently has zero
expectation:
$\mathbb{E}_{p(\mathbf{x}|\mathbf{y})}[\mathbf{g}_\theta] =
\mathbf{0}$.
A single tree thus produces control variates for all $k$ components
simultaneously, without requiring a separate network per component.

While a single tree produces valid control variates, it is limited by
its fixed binary subdivision pattern.
To improve coverage, we use $L$ independent ensemble members, each
with its own parameters and a random permutation
$\mathbf{P}_\ell$ applied to the input dimensions (outputs are
mapped back before computing $g_j$). Since the trace is invariant
under conjugation,
$\tr(\mathbf{P}_\ell\,\mathbf{J}\,\mathbf{P}_\ell^{-1}) =
\tr(\mathbf{J})$, each permuted tree still satisfies Stein's
identity (Appendix~\ref{app:stein}), and different permutations
capture complementary cross-dimension interactions.
The averaged control variate
\begin{equation}\label{eq:ensemble}
  \mathbf{g}_{\text{ens}}(\mathbf{x},\mathbf{y})
  = \frac{1}{L}\sum_{\ell=1}^{L}
    \mathbf{g}_{\theta_\ell}(\mathbf{x},\mathbf{y})
\end{equation}
preserves
$\mathbb{E}_{p(\mathbf{x}|\mathbf{y})}
[\mathbf{g}_{\text{ens}}(\mathbf{x},\mathbf{y})] = \mathbf{0}$;
we verify this empirically in
Appendix~\ref{app:stein} (Figure~\ref{fig:stein_verification}).
We ablate the effect of ensemble size in
Section~\ref{sec:ablation}.

\subsection{Training}\label{sec:training}

Our purpose is to find parameters $\theta$ that minimize
the total variance of the controlled estimator
$\mathbf{h} - \mathbf{g}$. Since
$\mathbb{E}_{p(\mathbf{x}|\mathbf{y})}
[\mathbf{g}_\theta(\mathbf{x},\mathbf{y})] = \mathbf{0}$
by Stein's identity, we have
\begin{align}\label{eq:mse_var}
  \mathbb{E}\bigl[\|\mathbf{h} - \mathbf{g}\|^2\bigr]
  &= \textstyle\sum_{j=1}^k \text{Var}(h_j - g_j)
  + \|\mathbb{E}[\mathbf{h}]\|^2,
\end{align}
where all expectations are conditional on the observation~$\mathbf{y}$
and the second term is constant in $\theta$. Expanding each
component gives
$\text{Var}(h_j - g_j) = \text{Var}(h_j) + \text{Var}(g_j) -
2\,\text{Cov}(h_j, g_j)$, so minimizing the MSE simultaneously
penalizes large $\text{Var}(g_j)$ and rewards high
$\text{Cov}(h_j, g_j)$---automatically balancing the two
contributions to variance reduction. This yields the training
objective:
\begin{align}\label{eq:loss}
  \{\theta_\ell^*\} = \arg\min_{\{\theta_\ell\}}
  \mathbb{E}\!\left[
      \left\|\mathbf{h}
      - \mathbf{g}_{\text{ens}}
      \right\|^2\right],
\end{align}
where the expectation is over $(\mathbf{x},\mathbf{y}) \sim p(\mathbf{x},\mathbf{y})$.
In practice, the expectation is approximated using $N$ training
samples $\{(\mathbf{x}_i, \mathbf{y}_i)\}_{i=1}^N$ generated by
sampling $\mathbf{x}_i \sim p_{\text{prior}}(\mathbf{x})$ and
computing
$\mathbf{y}_i = \mathcal{F}(\mathbf{x}_i) +
\boldsymbol{\epsilon}_i$. The score function
$\nabla_\mathbf{x} \log p(\mathbf{x}_i|\mathbf{y}_i)$ is evaluated
at each training pair using any of the sources described in
Section~\ref{sec:score}.

\subsection{Inference}

For a new observation $\mathbf{y}_{\text{obs}}$, we draw $M$
posterior samples using any available sampler (cf. line~9 of Algorithm~\ref{alg:cncv}) and compute the
variance-reduced estimate (line~15):
\begin{equation}\label{eq:inference}
  \widehat{\mathbb{E}}[\mathbf{h}(\mathbf{x})]
  = \frac{1}{M}\sum_{j=1}^M
  \bigl(\mathbf{h}(\mathbf{x}_j)
  - \mathbf{g}_{\text{ens}}(\mathbf{x}_j, \mathbf{y}_{\text{obs}})\bigr).
\end{equation}

\begin{algorithm}[t]
\caption{Conditional neural control variates (CNCV)}
\label{alg:cncv}
\begin{algorithmic}[1]
\REQUIRE Prior $p_{\text{prior}}$, forward model $\mathcal{F}$, score
  source, posterior sampler, ensemble size $L$, quantity of interest
  $\mathbf{h}$\vspace{-0.5em}
\STATE \textbf{// Offline training}
\FOR{$i = 1, \ldots, N_{\text{train}}$}
  \STATE Sample $\mathbf{x}_i \sim p_{\text{prior}}(\mathbf{x})$
  \STATE Compute $\mathbf{y}_i = \mathcal{F}(\mathbf{x}_i) +
    \boldsymbol{\epsilon}_i$
  \STATE Evaluate
    $\nabla_\mathbf{x} \log p(\mathbf{x}_i | \mathbf{y}_i)$ from
    chosen score source
\ENDFOR
\STATE Initialize $L$ hierarchical coupling layers with random input permutations
\STATE Optimize $\{\theta_\ell\}_{\ell=1}^L$ via
  equation~\eqref{eq:loss}
\STATE \textbf{// Online inference} (for any new
  $\mathbf{y}_{\text{obs}}$)
\STATE Draw $M$ posterior samples
  $\{\mathbf{x}_j\}_{j=1}^M \sim p(\mathbf{x}|\mathbf{y}_{\text{obs}})$
\FOR{$j = 1, \ldots, M$}
  \STATE Evaluate score
    $\nabla_\mathbf{x} \log p(\mathbf{x}_j | \mathbf{y}_{\text{obs}})$
  \STATE Evaluate
    $\mathbf{g}_{\text{ens}}(\mathbf{x}_j, \mathbf{y}_{\text{obs}})$ via
    equation~\eqref{eq:ensemble}
\ENDFOR
\RETURN $\widehat{\mathbb{E}}[\mathbf{h}] = \frac{1}{M}\sum_{j=1}^M
  \bigl(\mathbf{h}(\mathbf{x}_j) -
  \mathbf{g}_{\text{ens}}(\mathbf{x}_j, \mathbf{y}_{\text{obs}})\bigr)$
\end{algorithmic}
\end{algorithm}

The entire inference procedure involves only neural network forward
passes and score evaluations---$L$ coupling layer and score evaluations
per sample. When the dominant cost is the forward model---e.g., PDE
solves---this overhead is negligible. Crucially, no retraining is
required for new observations, maintaining full amortization.
Because the offline training cost is fixed while a per-observation
neural Stein control variate~\citep{BedaqueOh2024, Oh2025} must be
refit for every new observation, CNCV amortizes against the strongest
such baseline at approximately $190$ observations on the Gaussian
$d{=}4$ problem.

\section{Numerical experiments}\label{sec:experiments}

We validate the proposed CNCV framework on four inverse problems of
increasing complexity. We first consider three stylized
problems---Gaussian, Rosenbrock, and nonlinear---with tractable
posterior scores, then scale to a
PDE-constrained Darcy flow inverse problem
where the score must be learned from data.  Throughout, we report the
\emph{variance reduction factor}~(VRF)
$= \text{Var}(h - g) / \text{Var}(h)$, where lower values indicate
stronger variance reduction (VRF~$= 0$ is perfect, VRF~$= 1$ is no
improvement). In all experiments, $k = d$---i.e., the quantity of
interest has the same dimension as the unknown parameter. Code to
partially reproduce these results is available on
\href{https://github.com/luqigroup/cncv}{GitHub}.

\subsection{Stylized inverse problems}\label{sec:stylized}

These experiments quantify variance reduction and its effect on
estimation accuracy, study how performance scales with dimension and
sample size, ablate the ensemble size, and assess sensitivity to
learned scores. Full hyperparameters are in Appendix~\ref{app:hyperparams}.

\subsubsection{Gaussian posterior}

\begin{table}[t]
\centering
\caption{Variance reduction for Gaussian posteriors under per-dimension
hyperparameter tuning (recipes in Appendix~\ref{app:hyperparams}).
VRF $=$ Var$(h{-}g)$/Var$(h)$; lower is better. Values are mean $\pm$ std across
$d$ components, each averaged over 100 observations.}
\label{tab:gaussian_results}
\small
\begin{tabular}{ccc}
\toprule
$d$ & Mean Est.\ VRF & Var.\ Est.\ VRF \\
\midrule
$2$  & $0.009 \pm 0.000$ & $0.006 \pm 0.002$ \\
$4$  & $0.017 \pm 0.009$ & $0.008 \pm 0.003$ \\
$8$  & $0.034 \pm 0.010$ & $0.011 \pm 0.003$ \\
$16$  & $0.083 \pm 0.063$ & $0.120 \pm 0.102$ \\
\bottomrule
\end{tabular}
\end{table}

Consider the inverse problem
$\mathbf{y} = \mathbf{x} + \boldsymbol{\epsilon}$ with
$\mathbf{x} \in \mathbb{R}^d \sim \mathcal{N}(\mathbf{0},
\boldsymbol{\Sigma}_{\text{prior}})$ and
$\boldsymbol{\epsilon} \sim \mathcal{N}(\mathbf{0}, \sigma^2
\mathbf{I})$ with $\sigma = 0.3$. The prior covariance matrix is a randomly generated positive definite matrix. The analytical posterior is Gaussian with
known mean and covariance.
We evaluate (i)~posterior mean estimation with
$\mathbf{h}(\mathbf{x}) = \mathbf{x}$, and (ii)~posterior variance
estimation with
$\mathbf{h}(\mathbf{x}) = (\mathbf{x} - \boldsymbol{\mu}_{\text{post}})^2$.

\begin{figure}[t]
  \centering
  \includegraphics[width=\linewidth]{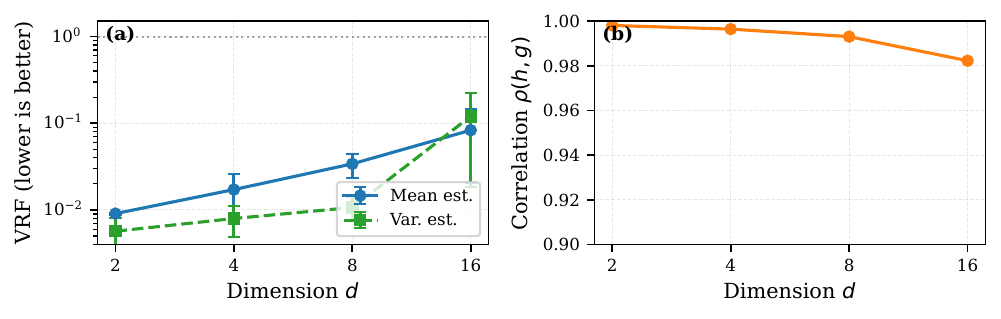}
  \caption{Dimension scaling of CNCV. (a)~VRF for mean
    and variance estimation across $d \in \{2,4,8,16\}$ (lower is
    better; $\text{VRF} = 1$ means no improvement).
    (b)~Correlation between $h$ and $g$ across dimensions.}
  \label{fig:dimension_scaling}
\end{figure}

\begin{figure}[t]
  \centering
  \includegraphics[width=\linewidth]{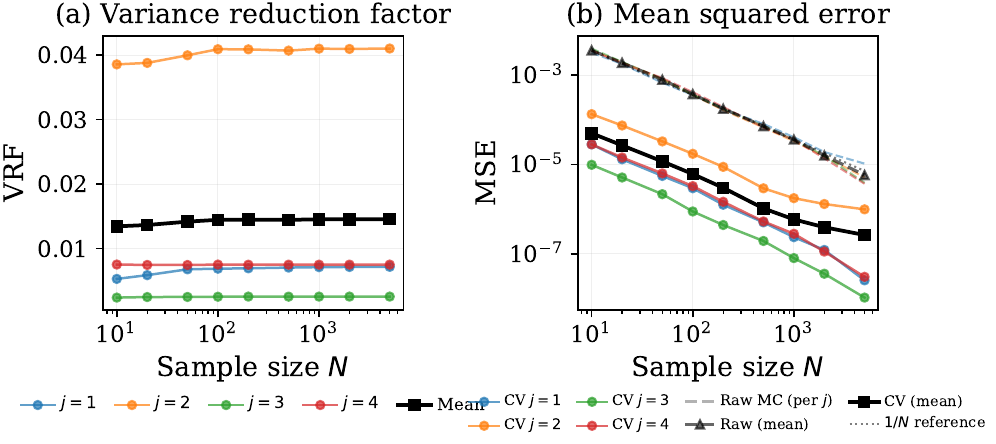}
  \caption{Sample efficiency for $d=4$ mean estimation. (a)~VRF is
    sample-size invariant at ${\sim}\,0.015$. (b)~MSE
    follows $1/N$ rate; the CV estimator provides a
    ${\sim}\,65\times$ effective sample increase.
    Index $j$ denotes the component.}
  \label{fig:sample_efficiency}
\end{figure}

Table~\ref{tab:gaussian_results} summarizes results
across dimensions. We observe that for mean estimation, CNCV achieves
VRF between $0.009$ and $0.083$ across $d \in \{2,4,8,16\}$, while for
variance estimation, VRF remains below $0.02$ at $d \le 8$ and
reaches $0.12$ at $d=16$ (each entry is
averaged over 100 test observations).
Figure~\ref{fig:dimension_scaling} shows that VRF grows mildly
with~$d$, while correlation between $h$ and $g$ remains above 0.98
for mean estimation. Figure~\ref{fig:sample_efficiency} shows the sample efficiency for
$d=4$. We observe from panel~(a) that the VRF is sample-size
invariant at ${\sim}\,0.015$ across sample sizes from 10 to 5k.
Panel~(b) shows MSE vs.\ sample size: while both estimators follow
the $1/N$ rate, the CV estimator has a substantially lower constant
factor, a ${\sim}\,65\times$ effective sample-size increase.
On this Gaussian target the posterior score is linear, so a classical
polynomial Stein control variate~\citep{MiraSolgiImparato2013} is
already near-optimal and can match CNCV on a single observation; this
is the easy special case. The moment the posterior departs from
Gaussianity the optimal control variate becomes nonlinear in the
score, and the polynomial baseline degrades sharply: under a
heavy-tailed Student-$t$ likelihood it increases variance on $95\%$ of
held-out observations while CNCV reduces it on every one
(Appendix~\ref{app:studentt}). The expressive coupling-layer ensemble
is therefore essential precisely on the non-Gaussian posteriors that
motivate this work.

\subsubsection{Rosenbrock posterior}

To assess CNCV on non-Gaussian posteriors, we consider the Rosenbrock
distribution---a benchmark with a strongly curved, banana-shaped
density, with prior
$p(\mathbf{x}) \propto \exp(-a(x_1 - \mu)^2 - b(x_2 - x_1^2)^2)$
where $\mu=0$, $a=0.5$, $b=1$. Combined with Gaussian likelihood
$p(\mathbf{y}|\mathbf{x}) = \mathcal{N}(\mathbf{x}, \sigma^2\mathbf{I})$
with $\sigma = 0.3$,
the posterior is non-Gaussian and exhibits strong nonlinear correlations
between components.

\begin{figure}[t]
  \centering
  \includegraphics[width=\linewidth]{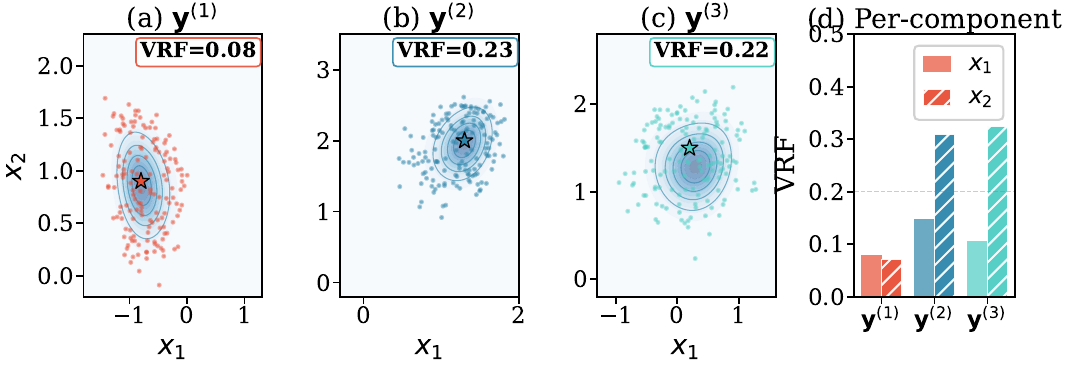}
  \caption{Amortized CNCV on Rosenbrock posterior. (a--c)~Test
    observations from left tail, right tail, and ridge (stars) yield
    diverse posteriors; VRF values are shown in each panel.
    (d)~Per-component VRF for each observation; the same trained model achieves VRF
    $\in [0.08, 0.23]$ across all cases (lower is better).}
  \label{fig:rosenbrock}
\end{figure}

Figure~\ref{fig:rosenbrock} demonstrates amortization across three
test observations from diverse regions: $\mathbf{y}^{(1)}$ from the
left tail, $\mathbf{y}^{(2)}$ from the right tail at high $x_2$, and
$\mathbf{y}^{(3)}$ near the ridge. Each observation (marked by stars)
induces a differently positioned banana-shaped posterior. Despite this
diversity, the \emph{same trained model} achieves VRF between
$0.08$ and $0.23$ across all cases (panel~d). This confirms that the control variate generalizes to any new observation.

\begin{figure}[t]
  \centering
  \includegraphics[width=\linewidth]{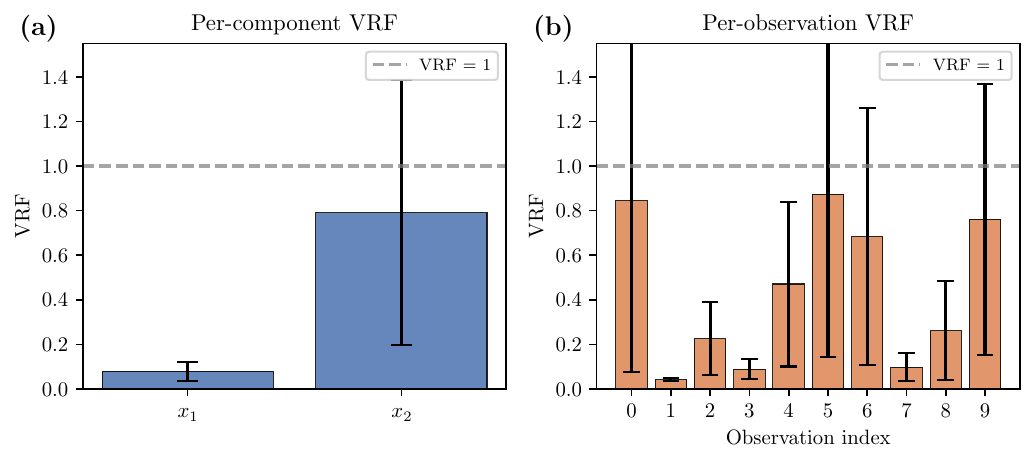}
  \caption{Posterior variance estimation on the Rosenbrock problem.
    (a)~Per-component VRF: $x_1$ achieves strong reduction while
    $x_2$ (banana direction) is harder.
    (b)~Per-observation VRF across 10 test observations (mean $0.44$).}
  \label{fig:rosenbrock_var}
\end{figure}
Aside from posterior mean estimation, CNCV also reduces variance for
posterior variance estimation.
Figure~\ref{fig:rosenbrock_var} uses
$\mathbf{h}(\mathbf{x}) = (\mathbf{x} -
\boldsymbol{\mu}_{\text{est}})^2$, where
$\boldsymbol{\mu}_{\text{est}}$ and the score are obtained
from a pretrained CNF (Section~\ref{sec:score}).
The first component ($x_1$) achieves $\text{VRF} = 0.08$, while the
banana direction ($x_2$)---where the posterior exhibits the strongest
nonlinear correlations---is harder ($\text{VRF} = 0.79$), yielding a
mean VRF of $0.44$ across 10 test observations.

\subsubsection{Nonlinear forward model}

\begin{figure}[t]
  \centering
  \includegraphics[width=0.925\linewidth]{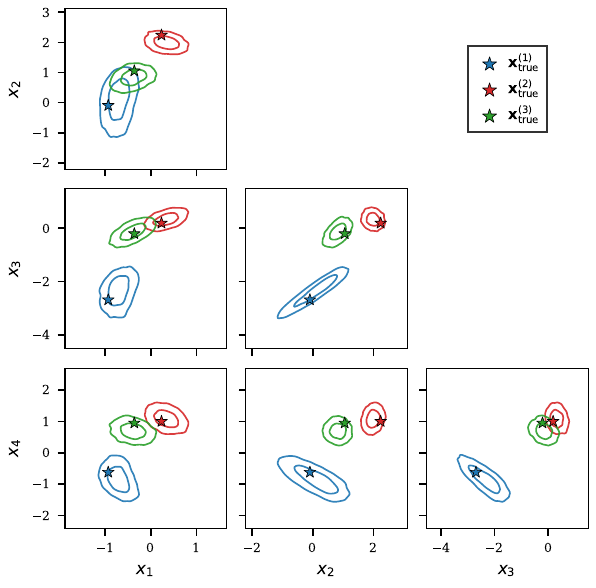}
  \caption{Corner plot of the nonlinear posterior ($d=4$) for three
    test observations (blue, red, green). Contours show 50\% and 90\%
    HDRs; stars mark the true parameters.}
  \label{fig:nonlinear_posterior}
\end{figure}

\begin{figure}[t]
  \centering
  \includegraphics[width=\linewidth]{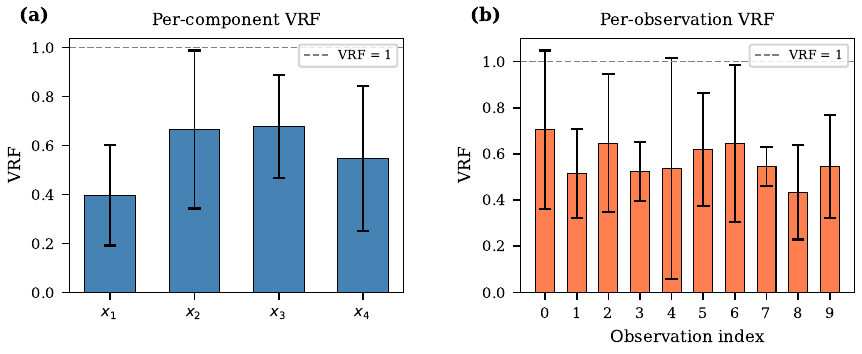}
  \caption{CNCV variance reduction on the nonlinear forward model
    ($d=4$). (a)~Per-component VRF averaged over 10 test
    observations. (b)~Per-observation VRF (mean $0.57$).}
  \label{fig:nonlinear}
\end{figure}

We next consider the nonlinear forward model
$\mathcal{F}(\mathbf{x}) = \mathbf{A}\mathbf{x} + \sin(\mathbf{x})$
in $d = 4$, where $\mathbf{A}$ has condition number~2.
Figure~\ref{fig:nonlinear_posterior} shows the non-Gaussian posteriors
for three test observations. At test time, posterior samples come from
a denoising diffusion probabilistic
model~\citep{HoJainAbbeel2020} trained on the same joint
distribution, providing fast amortized sampling.
CNCV achieves $\text{VRF} = 0.57 \pm 0.29$ across 10 test
observations.
Figure~\ref{fig:nonlinear} shows the per-component and
per-observation VRFs.

\subsubsection{Training efficiency and ensemble size}\label{sec:ablation}

\begin{figure}[t]
  \centering
  \includegraphics[width=\linewidth]{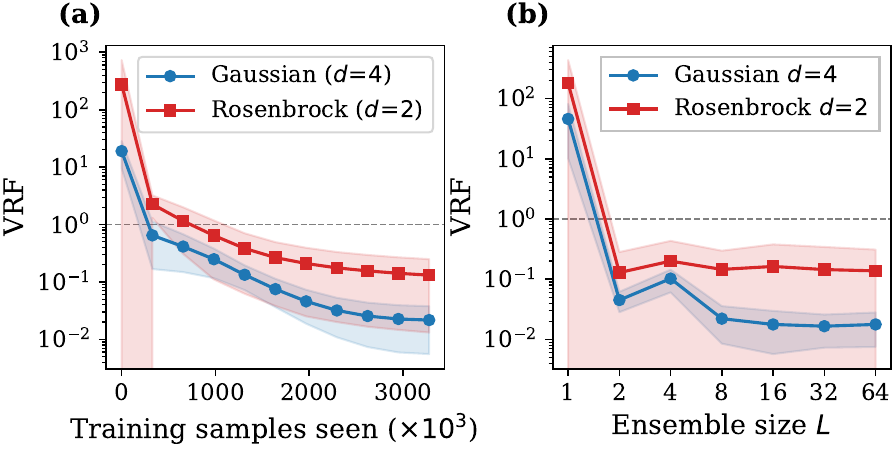}
  \caption{(a)~VRF versus total training samples seen.
    (b)~VRF versus ensemble size~$L$.
    Shading shows $\pm 1$ standard deviation across test
    observations. Both panels show Gaussian $d\!=\!4$ and
    Rosenbrock $d\!=\!2$ mean estimation, using the reference
    configuration to isolate training dynamics; the
    per-dimension-tuned results are in
    Table~\ref{tab:gaussian_results}.}
  \label{fig:training_ablation}
\end{figure}

Figure~\ref{fig:training_ablation}(a) shows VRF versus total
training samples seen for both the Gaussian ($d=4$) and Rosenbrock
($d=2$) problems. We observe that both converge rapidly: the Gaussian
VRF drops from $\sim$19 to below 0.03, and the Rosenbrock from
$\sim$277 to below 0.14, within the first few passes over the
dataset. Figure~\ref{fig:training_ablation}(b) shows VRF as a
function of the ensemble size~$L$. A single member ($L=1$) yields
almost no variance reduction; however, two members already reduce VRF
by roughly three orders of magnitude on the Gaussian problem, and performance
saturates beyond $L\approx 8$. This motivates the default $L=16$ used
throughout our experiments.

\subsubsection{Sensitivity to learned scores}

\begin{table}[t]
\centering
\caption{Analytical vs.\ CNF-learned score comparison.
  VRF = Var$(h{-}g)$/Var$(h)$; lower is better.
  Gaussian and nonlinear: mean $\pm$ std across components;
  Rosenbrock: range over three test observations.
  Each ensemble is trained and evaluated with the same score source.
  The Gaussian architecture is held fixed at the reference configuration
  ($64$ hidden units, $3$-layer MLP) to isolate the score-source effect;
  the per-dimension-tuned results appear in Table~\ref{tab:gaussian_results}.}
\label{tab:cnf_score_comparison}
\footnotesize
\setlength{\tabcolsep}{3.5pt}
\begin{tabular}{@{}llccc@{}}
\toprule
Problem & Score & VRF & Red.\ (\%) & $\rho$ \\
\midrule
Gauss.\ $d\!=\!4$   & Analyt.\ & $0.040 \pm 0.011$ & 96     & ${>}0.99$ \\
Gauss.\ $d\!=\!4$   & CNF      & $0.058 \pm 0.002$ & 94     & $0.98$   \\
\midrule
Rosenb.\ $d\!=\!2$  & Analyt.\ & $0.08$--$0.23$    & 77--92 & ---       \\
Rosenb.\ $d\!=\!2$  & CNF      & $0.010 \pm 0.002$ & 99     & ${>}0.99$ \\
\midrule
Nonlin.\ $d\!=\!4$  & Analyt.\ & $0.57 \pm 0.29$   & 43     & ---       \\
Nonlin.\ $d\!=\!4$  & CNF      & $0.57 \pm 0.26$   & 43     & ---       \\
\bottomrule
\end{tabular}
\end{table}

\begin{figure}[t]
  \centering
  \includegraphics[width=\linewidth]{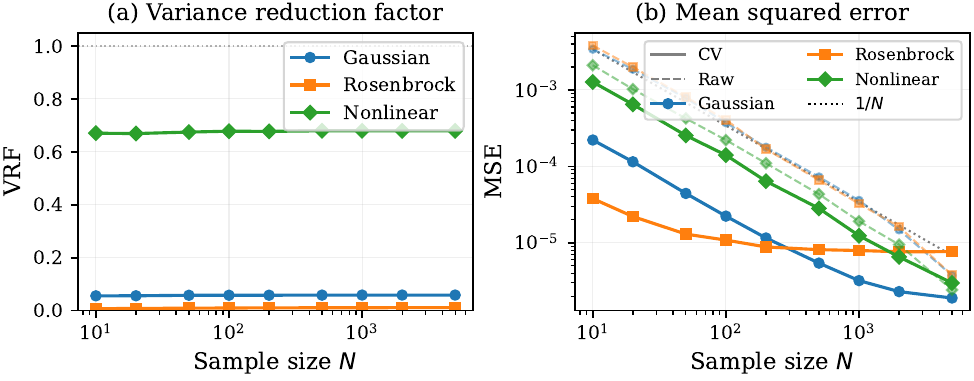}
  \caption{Sample efficiency with CNF-learned scores.
    (a)~VRF is sample-size invariant.
    (b)~The CV estimator (solid) provides a constant-factor MSE
    improvement over raw MC (dashed). MSE plateaus at large~$N$
    due to residual bias from
    the CNF score approximation.
    }
  \label{fig:cnf_efficiency}
\end{figure}

All preceding experiments used posterior scores computed directly from
the known likelihood and prior.
In practice, however, the score must often be learned from
data---particularly for inverse problems governed by expensive PDEs.
To assess robustness to score approximation error, we train a
CNF to approximate
$p(\mathbf{x}|\mathbf{y})$ via negative log-likelihood on joint
samples drawn from $p(\mathbf{x},\mathbf{y})$. The score is then
obtained by automatic differentiation of the change-of-variables
log-density (CNF architecture details in
Appendix~\ref{app:hyperparams}). As discussed in
Section~\ref{sec:score}, using the CNF score together with CNF
samples yields an estimator unbiased for
$\mathbb{E}_{p_\varphi}[h(\mathbf{x})]$---i.e., expectations under
the learned posterior rather than the true posterior. MSE in
Figure~\ref{fig:cnf_efficiency} is accordingly computed against the
CNF sample mean.
Table~\ref{tab:cnf_score_comparison} compares analytical and
CNF-learned scores on all three benchmarks. On the Gaussian problem
($d\!=\!4$), the CNF score increases VRF from $0.040$ to
$0.058$---a modest degradation consistent with score approximation
error. On the Rosenbrock problem ($d\!=\!2$), the CNF score achieves
VRF of $0.010 \pm 0.002$ with $\rho > 0.99$, lower than the
analytical-score VRF ($0.08$--$0.23$)---likely because the CNF
provides i.i.d.\ samples, whereas the analytical-score experiment
relies on correlated MCMC draws. On the nonlinear problem
($d\!=\!4$), the CNF
score achieves $\text{VRF} = 0.57 \pm 0.26$, virtually identical to
the analytical score ($0.57 \pm 0.29$), despite 17.5\% relative score
error. These results suggest that the coupling layer architecture is
robust to moderate score approximation errors.
Appendix~\ref{app:score_pert} confirms this directly: with the trained
ensemble held fixed and the analytical Gaussian score perturbed by
controlled noise, the VRF degrades gracefully and monotonically,
rising only to $0.35$ even at a $57\%$ relative score error---well
beyond the error of the learned scores used here.
Figure~\ref{fig:cnf_efficiency} confirms that VRF remains sample-size
invariant with CNF-learned scores, and the CV estimator maintains a
constant-factor MSE improvement over raw MC across all sample sizes.

\begin{figure}[t]
  \centering
  \includegraphics[width=\linewidth]{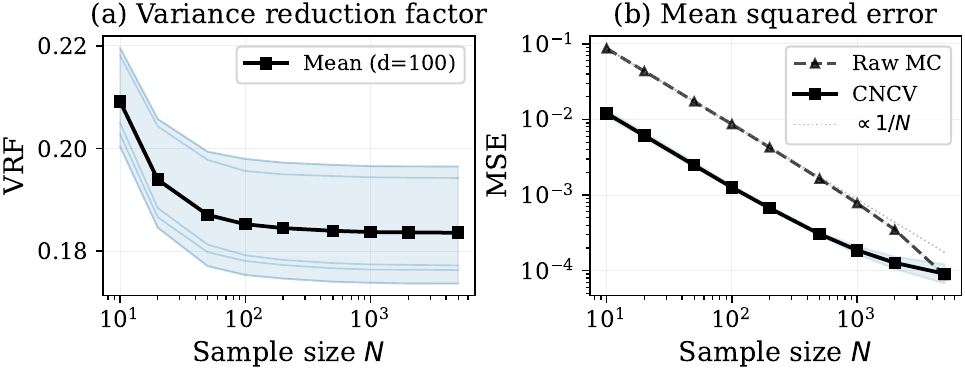}
  \caption{Sample efficiency for Darcy flow.
    (a)~VRF is sample-size invariant at ${\sim}\,0.18$.
    (b)~MSE of the posterior mean estimator follows $1/N$ for both
    raw MC and CNCV, with the CV estimator achieving a
    ${\sim}\,5.5\times$ constant-factor improvement. At large~$N$,
    both curves plateau because the reference mean is estimated from
    a finite CNF sample pool.}
  \label{fig:darcy_sample_efficiency}
\end{figure}

\subsection{PDE-constrained inverse problem}\label{sec:darcy}

\begin{figure*}[!t]
  \centering
  \includegraphics[width=\textwidth]{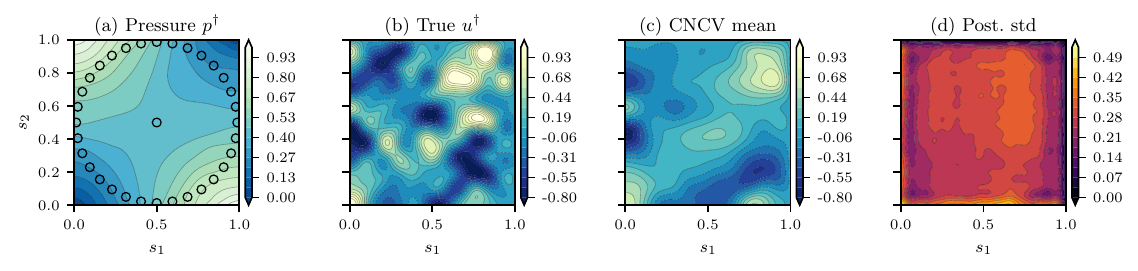}
  \vspace{-3em}
  \caption{Darcy flow posterior summary.
    (a)~Forward PDE pressure $p^\dagger$ with sensor locations
    (open circles);
    (b)~true log-permeability $u^\dagger$;
    (c)~CNCV posterior mean from $N=5{,}000$ CNF samples;
    (d)~pointwise posterior standard deviation.
    All fields reconstructed at $128 \times 128$ from KL coefficients.}
  \label{fig:darcy_posterior}
\end{figure*}

\begin{figure*}[!t]
  \centering
  \includegraphics[width=\textwidth]{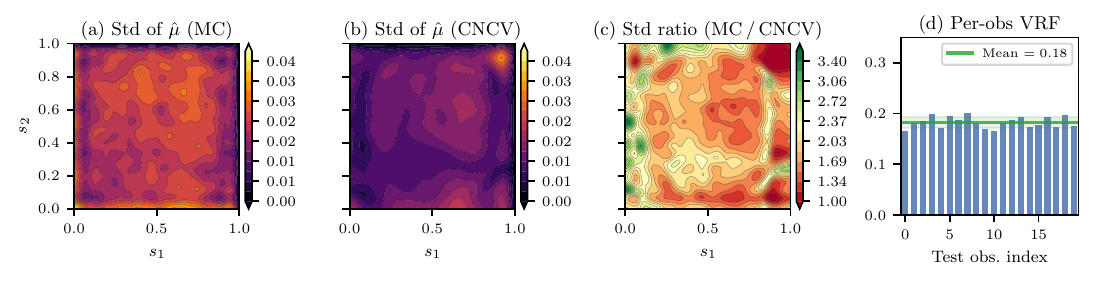}
  \vspace{-3em}
  \caption{Variance of the mean estimator on the Darcy flow problem.
    (a,~b)~Pixelwise standard deviation of the posterior mean
    estimator $\hat{\boldsymbol\mu} = \frac{1}{N}\sum_i \mathbf{x}_i$
    for vanilla MC and CNCV;
    (c)~std ratio (MC\,/\,CNCV), where values ${>}\,1$ indicate
    that CNCV reduces estimator variance;
    (d)~per-observation VRF across 20 held-out test observations.}
  \label{fig:darcy_estimator_std}
\end{figure*}

We now apply CNCV to a higher-dimensional, PDE-constrained problem:
estimating the log-permeability field in a groundwater flow model
governed by Darcy's equation~\citep{DarcySeminal1856},
$-\nabla \cdot (e^{u(\mathbf{s})}\,\nabla p(\mathbf{s})) = 0$ on
$\mathcal{D} = [0,1]^2$, with mixed boundary
conditions~\citep{BeskosEtAl2017}. The unknown log-permeability
$u(\mathbf{s})$ is observed through $m=33$ pressure sensors with
Gaussian noise ($\sigma_y = 0.01$). The PDE forward model costs
${\sim}\,0.5$\,s per solve, the posterior is high-dimensional, and the
score must be learned from data.
Following~\citet{BeskosEtAl2017}, the prior on $u$ is a Gaussian random field (GRF) represented via a cosine Karhunen--Lo\`eve~(KL) expansion with truncation $K=10$, giving $u(\mathbf{s}) = \sum_{j=1}^{100} z_j \sqrt{\lambda_j}\,\psi_j(\mathbf{s})$ where $\lambda_j$ are the GRF eigenvalues and $\psi_j$ are cosine eigenfunctions. The training data consists of $d = K^2 = 100$ KL coefficients $\mathbf{z} \in \mathbb{R}^{100}$ obtained by expanding GRF samples, paired with the corresponding observations. The forward model reconstructs the field from $\mathbf{z}$, solves the PDE on a $40 \times 40$ grid using Devito~\citep{devito-api, devito-compiler}, and extracts pressures at the sensor locations.
Since the posterior score is not available in closed form for this
nonlinear PDE, we train a CNF on
120k joint $(\mathbf{z}, \mathbf{y})$ samples to serve as both
posterior sampler and score source
(Appendix~\ref{app:cnf_quality} verifies the quality of the learned
score). The CNCV ensemble uses $L=16$ depth-3 trees;
full hyperparameters are in Appendix~\ref{app:hyperparams}.

Figure~\ref{fig:darcy_sample_efficiency} confirms that the VRF is
sample-size invariant at ${\sim}\,0.18$, with the CV estimator
achieving ${\sim}\,5.5\times$ lower MSE---an effective $5.5\times$ sample
size increase on this 100-dimensional problem. Because CNCV draws its
samples from the trained CNF rather than solving the PDE, this gain is
effectively free at inference: producing an estimate takes
${\sim}\,0.5$\,s, against ${\sim}\,100$\,s for the 200 PDE solves a
raw Monte-Carlo estimate of comparable accuracy would require
(Appendix~\ref{app:hyperparams}). Figure~\ref{fig:darcy_posterior} shows the posterior summary for a
held-out observation: the CNCV mean closely matches the truth
(Appendix~\ref{app:cnf_bias} quantifies the CNF posterior-mean bias
against held-out truth), with
posterior uncertainty concentrated in the domain interior---away from
sensors and Dirichlet boundaries. Since the reconstruction is linear
in the KL coefficients, the posterior standard deviation is governed
by the sensor geometry rather than the specific observation values,
and is therefore consistent across test cases
(Appendix~\ref{app:darcy_gallery}).
As observed from Figure~\ref{fig:darcy_estimator_std}, CNCV reduces
estimator standard deviation most strongly along large-scale
structures; per-observation VRF is tightly concentrated around $0.18$
across 20 test observations. To confirm that CNCV is not specific to
Darcy flow, we apply it to a second, physically distinct
PDE-constrained problem---full-waveform inversion governed by the
Helmholtz equation~\citep{VirieuxOperto2009}---where it attains a
per-observation VRF of $0.27 \pm 0.03$, below unity on all $20$
held-out observations (Appendix~\ref{app:helmholtz}).

\section{Related work}\label{sec:related}

Stein-based control variates have a rich history in MC variance
reduction. The zero-variance principle of
\citet{AssarafCaffarel1999} inspired kernel-based control
functionals~\citep{OatesGirolamiChopin2017, OatesCockayne2019,
  SouthOatesKarvonenEtAl2022} and polynomial/regularized MCMC
estimators~\citep{MiraSolgiImparato2013,
  SouthOatesMiraDrovandi2023}, while \citet{Si2022} unified
polynomial, kernel, and neural function classes under a scalable Stein
objective. Neural networks have also been used to parameterize control
variates outside the Stein framework---for gradient
estimation~\citep{Grathwohl2018}, rendering~\citep{Muller2020},
general MC integration~\citep{Wan2020}, and vector-valued~\citep{Sun2023vv}
and meta-learning~\citep{Sun2023meta} settings---and within it, for
lattice field theory~\citep{BedaqueOh2024, Oh2025, HasanEtAl2024} and
reinforcement-learning policy training~\citep{li2026reverse}. However, all of
these methods are \emph{non-amortized}: each must be fitted for a
specific target distribution, requiring MCMC samples and an
$\mathcal{O}(n^3)$ kernel solve~\citep{OatesGirolamiChopin2017}, or
least-squares regression on MCMC
output~\citep{MiraSolgiImparato2013}.
In a direct head-to-head on the Gaussian $d{=}4$ problem, CNCV attains
the lowest variance reduction factor among these baselines, and on a
bimodal Gaussian mixture it continues to reduce variance as the
dimension grows to $d{=}64$, where the degree-one polynomial basis
saturates and the degree-two basis becomes intractable---matching a
per-observation neural Stein control variate while training only once
(Appendix~\ref{app:gmm}).

In contrast, amortized methods train a single model that generalizes
across observations~\citep{BleiEtAl2017}. Conditional normalizing
flows~\citep{RezendeMohamed2015, Papamakarios2021} and simulation-based
inference~\citep{Cranmer2020, Radev2020} learn density estimators
$p_\varphi(\mathbf{x}|\mathbf{y})$ from joint samples, with extensions
to inverse problems~\citep{SiahkoohiRizzutiHerrmann2022,
  SiahkoohiRizzutiOrozcoEtAl2023, OrozcoSiahkoohi2025}; importance sampling
can further reduce variance through better
proposals~\citep{Dax2023}, complementary to control variates. However,
to our knowledge, no prior work has proposed \emph{amortized control
variates}---observation-conditioned functions
$g(\mathbf{x}, \mathbf{y})$ with zero posterior expectation that
generalize to unseen $\mathbf{y}$ without retraining.

A complementary line of work reduces cost along the axis of
forward-model fidelity rather than the observation. Multifidelity and
multilevel Monte Carlo methods combine many cheap low-fidelity solves
with a few expensive high-fidelity ones to lower the cost of a fixed
estimator~\citep{PeherstorferWillcoxGunzburger2018}, and multilevel
delayed-acceptance MCMC~\citep{LykkegaardEtAl2023} accelerates
posterior sampling through a hierarchy of coarsened forward models.
These methods operate on an axis orthogonal to CNCV: multifidelity
reduces cost over forward-model fidelity, whereas CNCV reduces variance
over the observation~$\mathbf{y}$. They also differ in their
amortization profile---multilevel delayed acceptance is asymptotically
exact but non-amortized, running a fresh sampler per observation, while
CNCV trains once and generalizes across observations at the cost of
inheriting the bias of the learned score. For applications requiring
posteriors at more than roughly twenty distinct observations, the
amortized cost of CNCV is preferable; for a single high-precision
posterior, an asymptotically exact sampler such as multilevel delayed
acceptance is the better choice.

\section{Discussion}\label{sec:discussion}

The proposed method integrates seamlessly with simulation-based
inference techniques~\citep{Cranmer2020}, which already produce joint
samples $(\mathbf{x}_i, \mathbf{y}_i)$ and learned posterior
approximations from which the score is readily available. CNCV
training consumes exactly these existing artifacts, adding no extra
forward model evaluations, and the inference-time cost of evaluating
the ensemble of coupling layers is comparable to a single shallow
network evaluation. At the same time, CNCV also accommodates
physics-based scores computed directly from the likelihood and prior,
making it applicable to problems where high-fidelity forward
evaluations are available and accurate score computation is paramount.
This flexibility makes CNCV attractive across a wide range of
settings---from problems constrained by partial differential
equations, where each forward solve dominates the computational
budget and any variance reduction translates directly into
proportional savings, to lower-cost problems where the analytical
score is readily accessible.

It is worth noting that CNCV reduces the variance of posterior
expectations computed from a given set of samples, but does not alter
the samples themselves or improve the underlying posterior
approximation. While the quality of the control variate depends on
score accuracy, our normalizing flow-based experiments demonstrate that learned
scores yield comparable variance reduction to analytical scores.
Since Stein's identity requires the score to correspond to the
sampling distribution, it is crucial that the score and posterior
samples originate from the same source---e.g., both from a
CNF---to preserve the zero-mean property of
the control variate. Consequently, CNCV is exactly unbiased for
expectations under the learned posterior, and any discrepancy with the
true posterior is attributable to the learned score alone---the
control variate neither introduces nor amplifies it. On the Darcy
problem we quantify this gap directly: the systematic field-space bias
of the CNF posterior mean is negligible (${\sim}\,{-}10^{-3}$), and the
residual error is dominated by the irreducible posterior spread on
weakly-constrained modes rather than by any bias CNCV contributes
(Appendix~\ref{app:cnf_bias}). The method also yields little benefit when the
quantity of interest is already nearly constant under the
posterior---so that its variance is small to begin with---or is only
weakly correlated with the score. For Gaussian posteriors, whose score is linear,
polynomial control variates~\citep{MiraSolgiImparato2013} can match
CNCV for a single observation; the value of CNCV lies in non-Gaussian
settings, where amortization across observations is needed and the
optimal control variate is nonlinear in the score---as
Appendix~\ref{app:studentt} shows on a heavy-tailed Student-$t$
likelihood, on which a per-observation polynomial control variate
instead increases variance. As with any
amortized method, we expect a gap relative to observation-specific
optimization~\citep{OrozcoSiahkoohi2025}; CNCV is complementary to
such refinement strategies and can be combined with them to further
reduce estimator variance.

Beyond the inverse problems studied here, amortized, Stein-based
control variates apply to other domains where Monte Carlo estimation
is a bottleneck---e.g., variance reduction of policy gradient estimators
in reinforcement learning~\citep{Greensmith2004, li2026reverse}, scaling to
higher-dimensional fields in latent spaces learned by functional
autoencoders~\citep{Bunker2025}, and amortized uncertainty
quantification for digital
twins~\citep{Herrmann2023digitaltwin}.

\begin{contributions}
  AS conceived the project, developed the method, implemented the
  code, designed and ran all experiments, and wrote the paper.
  HO contributed to the theoretical framework for Stein-based control
  variates and provided guidance on the neural network
  parameterization.
\end{contributions}

\begin{acknowledgements}
    Ali Siahkoohi acknowledges support from the Institute for Artificial
    Intelligence at the University of Central Florida. Hyunwoo Oh was
    supported in part by the U.S.\ Department of Energy, Office of Nuclear
    Physics under Award Number DE-FG02-93ER40762.
\end{acknowledgements}

\bibliography{references}

\clearpage
\appendix

\section{Derivation of Stein's identity for posterior
  distributions}\label{app:stein}

We provide a self-contained derivation of Stein's identity
(equation~\eqref{eq:stein}) in the context of posterior distributions
arising from Bayesian inverse problems. The derivation relies on the
divergence theorem in $\mathbb{R}^d$ and the product rule for
divergences.

\subsection{Setup and regularity conditions}

Let $p(\mathbf{x}|\mathbf{y})$ denote the posterior density over
$\mathbf{x} \in \mathbb{R}^d$ for a fixed observation
$\mathbf{y} \in \mathbb{R}^m$, and let
$\boldsymbol{\phi}(\mathbf{x}, \mathbf{y}) \colon \mathbb{R}^d \times
\mathbb{R}^m \to \mathbb{R}^d$ be a smooth vector-valued function. We
require:
\begin{enumerate}
  \item $p(\mathbf{x}|\mathbf{y})$ is continuously differentiable and
    strictly positive on $\mathbb{R}^d$;
  \item $\boldsymbol{\phi}(\cdot, \mathbf{y})$ is continuously
    differentiable for each $\mathbf{y}$;
  \item The boundary condition
    $\lim_{\|\mathbf{x}\| \to \infty}
    \boldsymbol{\phi}(\mathbf{x}, \mathbf{y}) \,
    p(\mathbf{x}|\mathbf{y}) = \mathbf{0}$ holds, ensuring that the
    surface integral at infinity vanishes.
\end{enumerate}
The boundary condition is satisfied whenever $p(\mathbf{x}|\mathbf{y})$
decays sufficiently fast---e.g., when the prior is Gaussian or has
sub-Gaussian tails---and $\boldsymbol{\phi}$ grows at most
polynomially.

\subsection{Derivation via integration by parts}

Consider the integral
\begin{equation}\label{eq:app_integral}
  I = \int_{\mathbb{R}^d}
  \nabla_\mathbf{x} \cdot
  \bigl[\boldsymbol{\phi}(\mathbf{x}, \mathbf{y}) \,
    p(\mathbf{x}|\mathbf{y})\bigr] \, d\mathbf{x},
\end{equation}
where
$\nabla_\mathbf{x} \cdot \mathbf{v} = \sum_{i=1}^d \partial v_i /
\partial x_i$ denotes the divergence. By the divergence theorem applied
to the ball $B_R = \{\mathbf{x} : \|\mathbf{x}\| \leq R\}$,
\begin{equation*}
  \int_{B_R}
  \nabla_\mathbf{x} \cdot
  \bigl[\boldsymbol{\phi} \, p\bigr] \, d\mathbf{x}
  = \oint_{\partial B_R}
  \bigl[\boldsymbol{\phi} \, p\bigr] \cdot \hat{\mathbf{n}} \, dS,
\end{equation*}
where $\hat{\mathbf{n}}$ is the outward normal to the sphere
$\partial B_R$. By the boundary condition (3), the surface integral
vanishes as $R \to \infty$, giving
\begin{equation}\label{eq:app_vanish}
  I = \int_{\mathbb{R}^d}
  \nabla_\mathbf{x} \cdot
  \bigl[\boldsymbol{\phi}(\mathbf{x}, \mathbf{y}) \,
    p(\mathbf{x}|\mathbf{y})\bigr] \, d\mathbf{x} = 0.
\end{equation}

\subsection{Expanding via the product rule}

We now expand the divergence in the above expression using the product
rule. For each component
$i$:
\begin{equation*}
  \frac{\partial}{\partial x_i}
  \bigl[\phi_i(\mathbf{x}, \mathbf{y}) \,
    p(\mathbf{x}|\mathbf{y})\bigr]
  = \frac{\partial \phi_i}{\partial x_i} \, p(\mathbf{x}|\mathbf{y})
  + \phi_i(\mathbf{x}, \mathbf{y}) \,
  \frac{\partial p(\mathbf{x}|\mathbf{y})}{\partial x_i}.
\end{equation*}
Summing over $i = 1, \ldots, d$ and using the identity
$\nabla p / p = \nabla \log p$:
\begin{align*}
  \nabla_\mathbf{x} \cdot
  \bigl[\boldsymbol{\phi} \, p\bigr]
  &= \Bigl(\sum_{i=1}^d
    \frac{\partial \phi_i}{\partial x_i}\Bigr) p
  + \Bigl(\sum_{i=1}^d \phi_i \,
    \frac{\partial p}{\partial x_i}\Bigr) \\
  &= \bigl(\nabla_\mathbf{x} \cdot \boldsymbol{\phi}\bigr) \, p
  + \bigl(\boldsymbol{\phi} \cdot \nabla_\mathbf{x} \log p\bigr) \, p.
\end{align*}
Substituting into equation~\eqref{eq:app_vanish} and dividing by $p$
inside the expectation:
\begin{equation}\label{eq:app_stein_final}
  \mathbb{E}_{p(\mathbf{x}|\mathbf{y})}\!\left[
    \nabla_\mathbf{x} \cdot \boldsymbol{\phi}(\mathbf{x}, \mathbf{y})
    + \boldsymbol{\phi}(\mathbf{x}, \mathbf{y}) \cdot
    \nabla_\mathbf{x} \log p(\mathbf{x}|\mathbf{y})
  \right] = 0,
\end{equation}
which is exactly equation~\eqref{eq:stein}. This establishes that
$g(\mathbf{x}, \mathbf{y}) = \nabla_\mathbf{x} \cdot
\boldsymbol{\phi} + \boldsymbol{\phi} \cdot \nabla_\mathbf{x} \log
p(\mathbf{x}|\mathbf{y})$ has zero expectation under the posterior
for \emph{any} $\boldsymbol{\phi}$ satisfying the regularity
conditions.

\paragraph{Component-wise identity.}
Since the divergence theorem applies to each component of
$\boldsymbol{\phi}\,p$ independently, the identity also holds
\emph{per component} before summing:
\begin{equation*}
  \mathbb{E}_{p(\mathbf{x}|\mathbf{y})}\!\left[
    \frac{\partial \phi_j}{\partial x_j}
    + \phi_j \, \frac{\partial \log p(\mathbf{x}|\mathbf{y})}
    {\partial x_j}
  \right] = 0, \quad j = 1, \ldots, d.
\end{equation*}
This justifies the vector-valued control variate in
equation~\eqref{eq:component_cv}: each component $g_{\theta,j}$
independently has zero expectation, so
$\mathbb{E}[\mathbf{g}_\theta] = \mathbf{0}$ without requiring the
components to sum.

\subsection{Application to coupling layer architectures}

For our coupling layer parameterization
(equation~\eqref{eq:coupling}), the divergence takes the explicit
form~\eqref{eq:div}. The triangular structure means
$\nabla_\mathbf{x} \cdot \boldsymbol{\phi}_\theta$ involves only the
diagonal of the Jacobian, which equals $1$ for untransformed
components and $s_{\theta,i}(\mathbf{x}_1, \mathbf{y})$ for
transformed components. The resulting control variate
\begin{equation*}
  g_\theta(\mathbf{x}, \mathbf{y})
  = \underbrace{
    \sum_{i \in \mathbf{x}_2} s_{\theta,i}(\mathbf{x}_1, \mathbf{y})
    + d_1
  }_{\text{divergence (exact)}}
  + \underbrace{
    \vphantom{\sum_{i \in \mathbf{x}_2}}
    \boldsymbol{\phi}_\theta(\mathbf{x}, \mathbf{y}) \cdot
    \nabla_\mathbf{x} \log p(\mathbf{x}|\mathbf{y})
  }_{\text{score term}}
\end{equation*}
In the above expression, $d_1 = \dim(\mathbf{x}_1)$. This control
variate satisfies
$\mathbb{E}_{p(\mathbf{x}|\mathbf{y})}[g_\theta(\mathbf{x},
\mathbf{y})] = 0$ exactly, with no approximation in the divergence
computation. This unbiasedness guarantee holds regardless of the neural
network parameters $\theta$ (provided the boundary conditions are met),
and is preserved under ensemble averaging
(equation~\eqref{eq:ensemble}).

For the hierarchical coupling layer, the recursive binary tree
structure computes the Jacobian diagonal as
$\diag(\nabla_\mathbf{x} \boldsymbol{\phi}) = [\mathbf{d}_{\text{upper}}; \mathbf{s} \odot \mathbf{d}_{\text{lower}}]$,
where $\mathbf{d}_{\text{upper}}$ and $\mathbf{d}_{\text{lower}}$ are
the diagonals from the upper and lower subtrees, and $\mathbf{s}$ is
the scale vector from the coupling at the current node. At leaf nodes,
$\mathbf{d} = \mathbf{1}$. The divergence is then
$\tr(\nabla_\mathbf{x} \boldsymbol{\phi}) = \mathbf{1}^\top
\mathbf{d}$, which remains analytically computable and exact.

\subsection{Boundary conditions for posterior distributions in
  inverse problems}

We verify that condition~(3) holds in typical inverse problem settings.
For a Gaussian prior $p_{\text{prior}}(\mathbf{x}) =
\mathcal{N}(\boldsymbol{\mu}_0, \boldsymbol{\Sigma}_0)$ and Gaussian
likelihood $p(\mathbf{y}|\mathbf{x}) = \mathcal{N}(\mathbf{x}, \sigma^2\mathbf{I})$, the posterior satisfies
$p(\mathbf{x}|\mathbf{y}) \leq C \exp(-\frac{1}{2}(\mathbf{x} -
\boldsymbol{\mu}_0)^\top \boldsymbol{\Sigma}_0^{-1} (\mathbf{x} -
\boldsymbol{\mu}_0))$ for some $C > 0$, since the likelihood is
bounded above by~1. This Gaussian tail decay ensures
$\boldsymbol{\phi}(\mathbf{x}, \mathbf{y})\,p(\mathbf{x}|\mathbf{y})
\to \mathbf{0}$ as $\|\mathbf{x}\| \to \infty$ whenever
$\boldsymbol{\phi}$ grows at most polynomially---which holds for
neural networks with bounded (tanh, sigmoid) or polynomially growing
(ReLU) activations. For non-Gaussian priors such as the Rosenbrock
density $\exp(-a(x_1{-}\mu)^2 - b(x_2{-}x_1^2)^2)$, the
super-exponential decay in all directions similarly ensures the
boundary conditions hold.

\subsection{Empirical verification}

\begin{figure}[t]
  \centering
  \includegraphics[width=\linewidth]{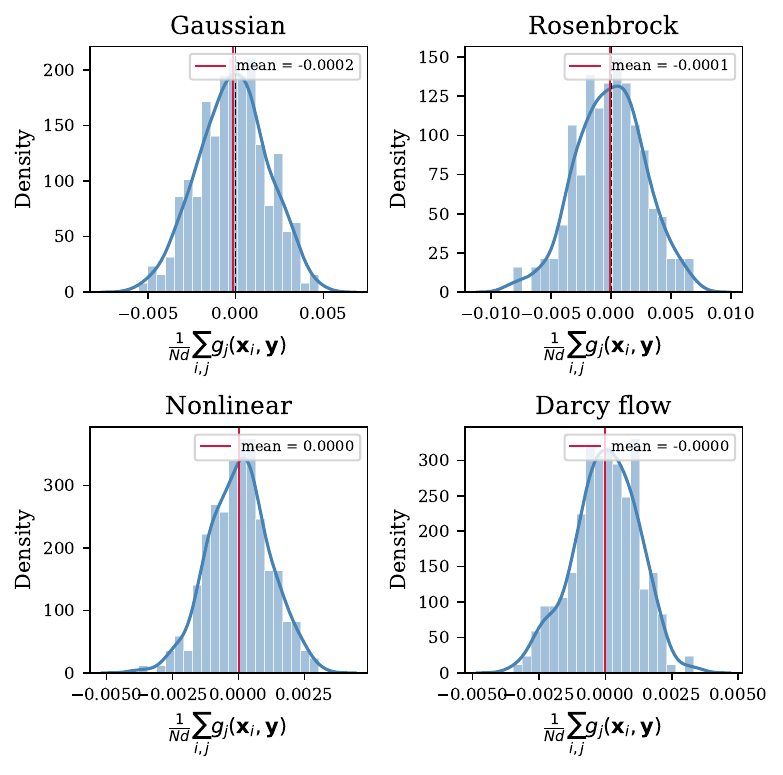}
  \caption{Empirical verification of Stein's identity.
    Each panel shows the distribution of
    $\frac{1}{Nd}\sum_{i,j} g_j(\mathbf{x}_i, \mathbf{y})$
    across 250 test observations ($N\!=\!5{,}000$ samples each).
    All four distributions are centered at zero, confirming the
    zero-mean guarantee.}
  \label{fig:stein_verification}
\end{figure}

Figure~\ref{fig:stein_verification} provides an empirical check of
Stein's identity. For each of 250 test observations~$\mathbf{y}$, we
draw $N = 5{,}000$ posterior samples and compute the dimension-averaged
sample mean of the control variate,
$\frac{1}{Nd}\sum_{i,j} g_j(\mathbf{x}_i, \mathbf{y})$, which should
be zero by construction. Across all four
experiments---Gaussian ($d\!=\!4$, analytical score), Rosenbrock
($d\!=\!2$), nonlinear ($d\!=\!4$), and Darcy ($d\!=\!100$, all three
with learned scores)---this quantity is tightly concentrated around zero
(mean~${\approx}\,0$, std~${\approx}\,10^{-3}$), confirming that the
zero-mean guarantee holds in practice.

\section{Per-component variance reduction on Darcy
  flow}\label{app:component_vrf}

\begin{figure}[t]
  \centering
  \includegraphics[width=\linewidth]{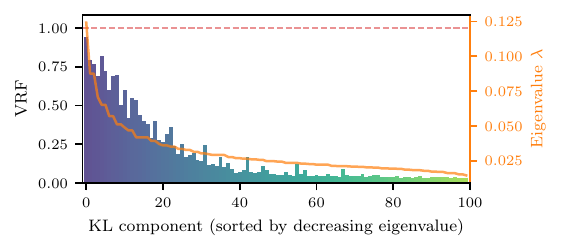}
  \caption{Per-KL-component VRF on the Darcy flow problem ($d=100$),
    sorted by decreasing prior eigenvalue $\lambda_j$ (orange curve,
    right axis). High-frequency modes (small $\lambda_j$, right side)
    achieve VRF as low as $0.04$; low-frequency modes
    (large $\lambda_j$, left side) have VRF~$\approx 1$.
    The dashed line marks VRF~$= 1$ (no improvement).}
  \label{fig:darcy_component_vrf}
\end{figure}

Figure~\ref{fig:darcy_component_vrf} shows the per-component VRF for
each of the 100 KL coefficients, sorted by decreasing prior
eigenvalue. Two regimes are visible:

\paragraph{Low-frequency modes (components 1--30).}
These modes correspond to large prior eigenvalues $\lambda_j$ and
carry most of the prior variance.
Because the data strongly constrains these components, the posterior
variance $\text{Var}(z_j|\mathbf{y})$ is already much smaller than
the prior variance $\lambda_j^2$. The control variate must match the
posterior mean function $\mathbb{E}[z_j|\mathbf{y}]$---a nonlinear
function of the observation---which is harder for the leading modes
where the data is most informative. This yields VRF close to~1.

\paragraph{High-frequency modes (components 60--100).}
These modes have small prior eigenvalues and are barely constrained
by the data: $p(z_j|\mathbf{y}) \approx p(z_j) = \mathcal{N}(0,1)$.
Here the posterior score is dominated by the prior score
$\nabla_{z_j}\log p(z_j) = -z_j$, which the coupling layers can
approximate easily, yielding VRF as low as~$0.04$.

The component-averaged VRF is approximately~$0.16$, while the
per-observation VRF reported in Section~\ref{sec:darcy}
(${\sim}\,0.18$) aggregates differently across samples.
In both cases, the overall reduction is substantial because the
majority of components (those with smaller eigenvalues) achieve very
low VRF.

\section{CNF quality diagnostics for Darcy
  flow}\label{app:cnf_quality}

Since the Darcy flow experiment relies entirely on a learned CNF for
both posterior sampling and score computation, we provide diagnostics
verifying the quality of the trained CNF
(Section~\ref{sec:darcy}).

\begin{figure}[t]
  \centering
  \includegraphics[width=\linewidth]{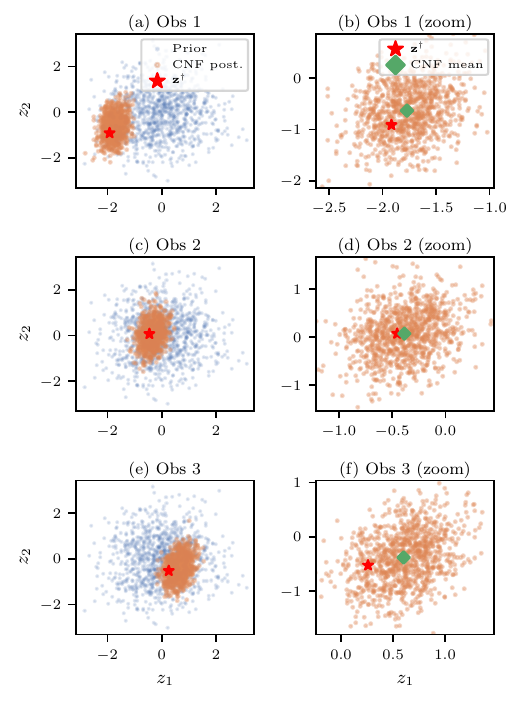}
  \caption{First two KL components of the CNF posterior for three
    held-out observations.
    Left column: prior samples (blue) and CNF posterior samples
    (orange) with the true parameter (red star); the CNF correctly
    concentrates around the truth.
    Right column: zoomed view showing the CNF posterior mean
    (green diamond) near the truth.}
  \label{fig:cnf_scatter}
\end{figure}

\begin{figure}[t]
  \centering
  \includegraphics[width=\linewidth]{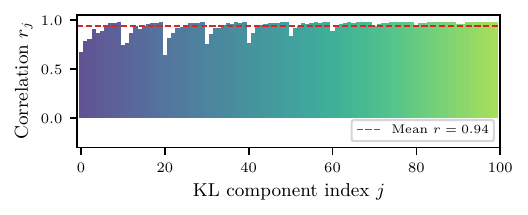}
  \caption{Per-component Pearson correlation between the CNF score
    $\hat s_j = \partial_{z_j}\log p_\varphi(\mathbf{z}|\mathbf{y})$
    and the physics-based posterior score $s_j$ (adjoint PDE + prior),
    evaluated at 100 held-out test observations.
    Mean $\bar{r} = 0.94$, median $0.97$.}
  \label{fig:cnf_score}
\end{figure}

\begin{figure}[t]
  \centering
  \includegraphics[width=\linewidth]{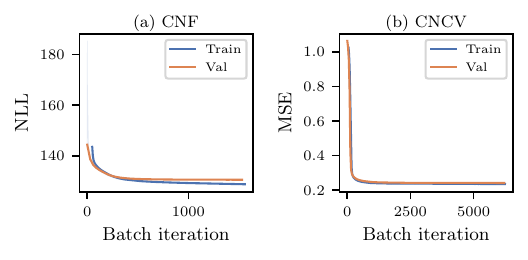}
  \caption{Training and validation loss curves.
    (a)~CNF negative log-likelihood: raw per-batch training loss
    (light), smoothed training loss (dark), and validation NLL
    per epoch.
    (b)~CNCV mean-squared error: same layout.
    Both models converge without overfitting.}
  \label{fig:loss_curves}
\end{figure}

Figure~\ref{fig:cnf_scatter} projects the 100-dimensional posterior
onto the first two KL components---i.e., those with the largest prior
eigenvalues---for three held-out observations. In each case, the CNF
posterior (orange) is correctly concentrated around the true parameter
(red star), with the CNF posterior mean (green diamond) close to the
truth. As expected, the prior samples (blue) are substantially more
dispersed, confirming that the CNF has learned meaningful posterior
concentration.

Figure~\ref{fig:cnf_score} compares the CNF score
$\nabla_{\mathbf{z}}\log p_\varphi(\mathbf{z}|\mathbf{y})$ against
the physics-based posterior score computed via adjoint PDE solves and
the GRF prior (Section~\ref{sec:darcy}), for each of the 100 KL
components across 100 held-out observations.  We observe that the
per-component Pearson correlation is high (mean $\bar{r} = 0.94$,
median $0.97$), confirming that the CNF learns an accurate score
across nearly all KL modes. We note that CNFs trained without weight
decay ($\lambda = 0$) produced inaccurate scores due to
coupling-scale explosion; weight decay of $10^{-4}$ on the AdamW
optimizer was essential for stable training.

Figure~\ref{fig:loss_curves} shows the training dynamics for both
the CNF and the CNCV ensemble. Both models converge smoothly, with
validation losses tracking the training losses closely, indicating no
significant overfitting.

\section{CNF posterior-mean bias on Darcy flow}\label{app:cnf_bias}

The Darcy flow experiment (Section~\ref{sec:darcy}) relies on a learned
CNF for both posterior sampling and the score. As established in
Section~\ref{sec:score}, the resulting estimator is unbiased for
expectations under the CNF posterior $p_\varphi$ rather than the true
posterior, and the control variate of Section~\ref{sec:cncv} is
zero-mean under $p_\varphi$ for any parameters---so it adds no bias
relative to it. The relevant question is then how closely the CNF
posterior mean tracks the truth, which we quantify against $20$
held-out ground-truth fields.

\begin{figure*}[t]
  \centering
  \includegraphics[width=0.92\textwidth]{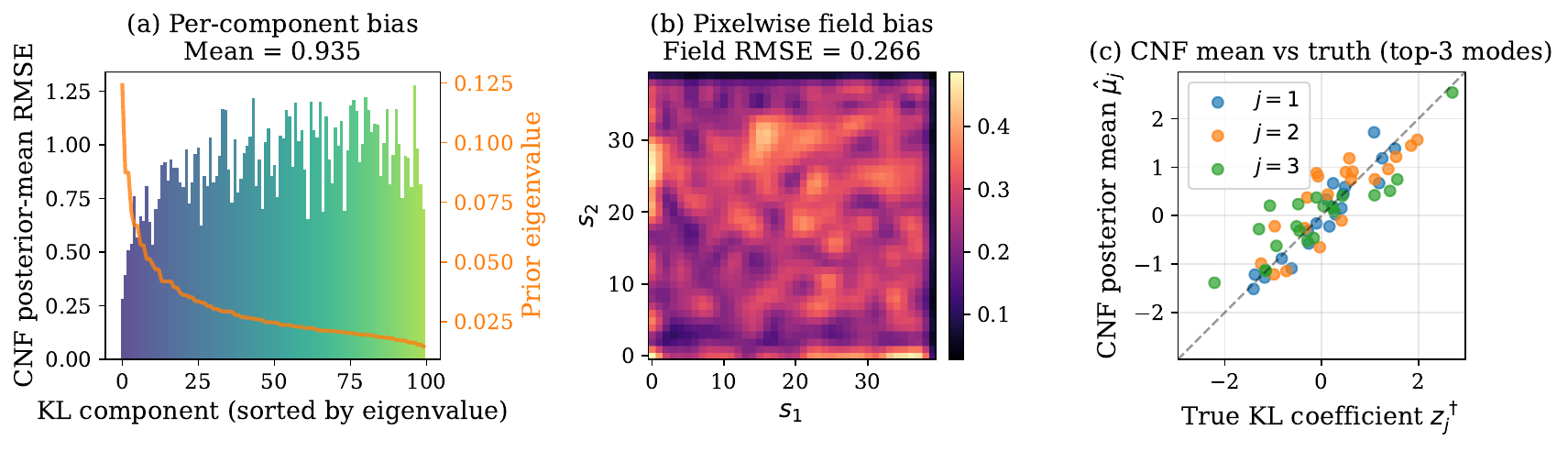}
  \caption{CNF posterior-mean discrepancy against $20$ held-out truths
    on the Darcy problem. The signed field-space bias is negligible
    (${\sim}\,-10^{-3}$); the unsigned errors---field-space RMSE
    (${\sim}\,0.27$) and the per-component KL root-mean-square error,
    which rises from ${\sim}\,0.28$ on the most data-informed mode
    toward ${\sim}\,1.0$ on weakly-constrained modes---reflect
    irreducible posterior spread (Appendix~\ref{app:component_vrf})
    rather than systematic bias.}
  \label{fig:cnf_bias}
\end{figure*}

Figure~\ref{fig:cnf_bias} reports the discrepancy between the CNF
posterior mean and the held-out truth, in both the field and KL
representations. The field-space bias, averaged over pixels and
observations, is negligible (${\sim}\,-10^{-3}$). The per-component KL
root-mean-square error rises from ${\sim}\,0.28$ on the most data-informed
mode (largest prior eigenvalue) toward ${\sim}\,1.0$ on the
weakly-constrained high-frequency modes. As in
Appendix~\ref{app:component_vrf}, the latter modes are barely informed
by the data, so their posterior approaches the unit-variance prior, and
a single posterior-mean estimate necessarily differs from any single
truth by an amount of order the prior standard deviation. The aggregate
KL error (${\sim}\,0.9$) is therefore dominated by this irreducible
posterior spread rather than by systematic bias. We emphasize that this
analysis concerns the CNF posterior relative to the true posterior;
quantifying or correcting that gap is distinct from the variance
reduction CNCV provides, which is exact with respect to the CNF
posterior.

\section{Additional Darcy posterior
  visualizations}\label{app:darcy_gallery}

\begin{figure*}[t]
  \centering
  \includegraphics[width=\textwidth]{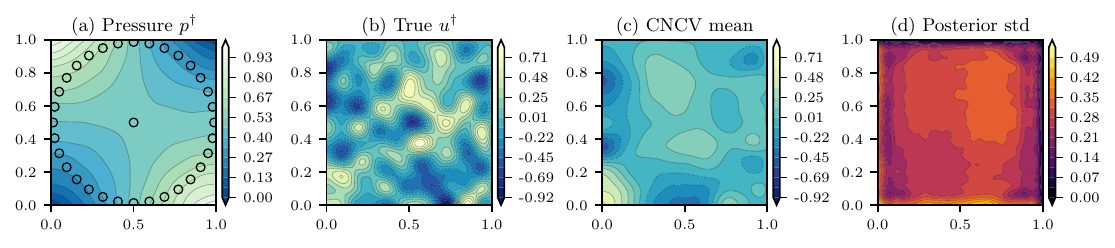}\\[-2pt]
  \includegraphics[width=\textwidth]{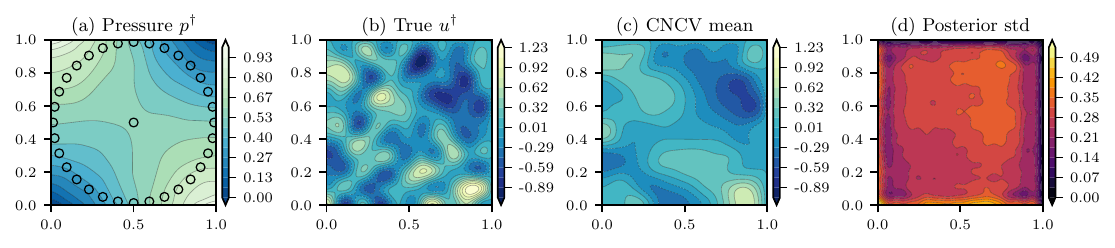}\\[-2pt]
  \includegraphics[width=\textwidth]{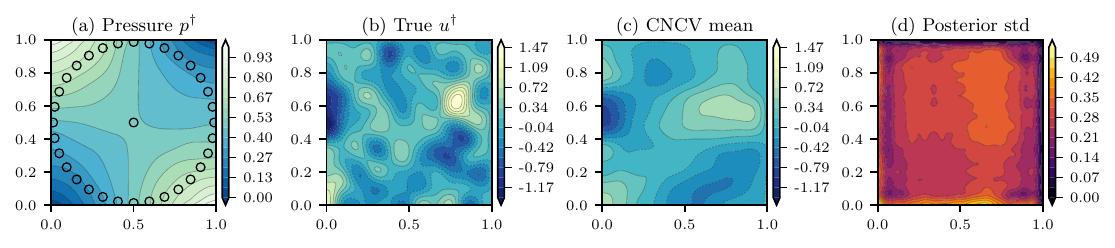}
  \caption{Darcy posterior summaries for three additional held-out test
    observations. Each row shows a different observation. Columns:
    (a)~pressure $p^\dagger$ with sensors,
    (b)~true $u^\dagger$,
    (c)~CNCV posterior mean,
    (d)~posterior std.}
  \label{fig:darcy_gallery}
\end{figure*}

\begin{figure*}[t]
  \centering
  \includegraphics[width=\textwidth]{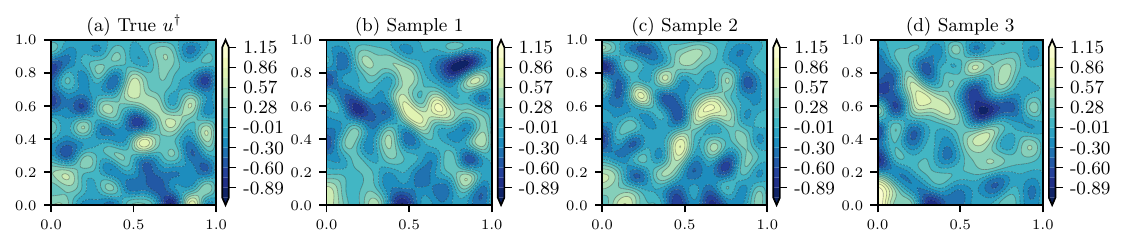}\\[-2pt]
  \includegraphics[width=\textwidth]{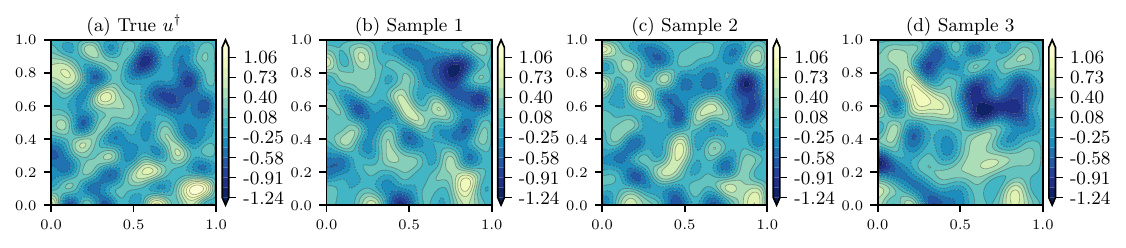}\\[-2pt]
  \includegraphics[width=\textwidth]{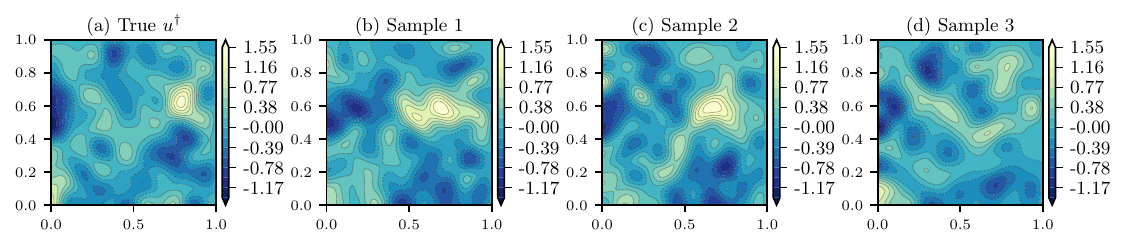}
  \caption{Individual posterior samples for the same three observations
    as Figure~\ref{fig:darcy_gallery}. Each row shows the true field
    (a) and three independent draws from the CNF approximate posterior
    (b--d). Samples preserve the frequency content of the truth; the
    smoothness of the posterior mean in Figure~\ref{fig:darcy_gallery}(c)
    arises from averaging over uncertain high-frequency modes.}
  \label{fig:darcy_posterior_samples}
\end{figure*}

Figure~\ref{fig:darcy_gallery} shows posterior summaries for three
additional held-out observations, demonstrating that CNCV generalizes
beyond the single observation shown in
Figure~\ref{fig:darcy_posterior}. Each row follows the same format:
pressure field $p^\dagger$ with sensor locations, true
log-permeability $u^\dagger$, CNCV posterior mean, and posterior standard
deviation. Across all observations, the CNCV posterior mean recovers the
large-scale structure of the truth, with posterior uncertainty concentrated in
the domain interior, away from sensors and Dirichlet boundaries.
Figure~\ref{fig:darcy_posterior_samples} shows individual posterior draws
for the same three observations; the samples retain the high-frequency
structure that averaging into the posterior mean smooths out.

\section{Helmholtz full-waveform inversion}\label{app:helmholtz}

To test whether CNCV transfers beyond Darcy flow, we apply it to a
second, physically distinct PDE-constrained inverse problem:
full-waveform inversion (FWI) governed by the Helmholtz equation, a
canonical Bayesian inverse problem in subsurface
geophysics~\citep{VirieuxOperto2009, BeskosEtAl2017}. The unknown is a
two-dimensional wave-speed perturbation on a $32\times32$ grid,
parameterized by the same $K=10$ cosine Karhunen--Lo\`eve basis as the
Darcy experiment ($d=100$); the observation~$\mathbf{y}$ stacks the
complex wavefields recorded at $40$ receivers from $5$ sources at
$4$~Hz, with additive noise. As in the Darcy experiment, the posterior
score is provided by a conditional normalizing flow trained on joint
$(\mathbf{x},\mathbf{y})$ samples, and CNCV is trained on the same
samples using that learned score.

\begin{figure*}[t]
  \centering
  \includegraphics[width=0.92\textwidth]{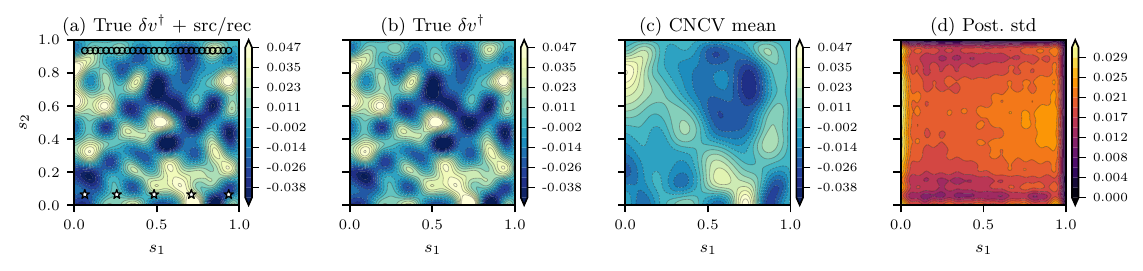}
  \caption{Helmholtz FWI posterior summary for a held-out observation.
    (a)~true wave-speed perturbation $\delta v^\dagger$ with the $5$
    sources (stars) and $40$ receivers (circles);
    (b)~true $\delta v^\dagger$;
    (c)~CNCV posterior mean;
    (d)~posterior standard deviation.}
  \label{fig:helmholtz_posterior}
\end{figure*}

\begin{figure}[t]
  \centering
  \includegraphics[width=\linewidth]{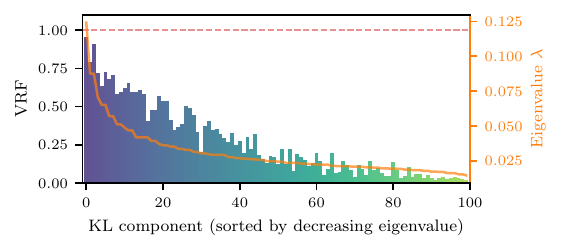}
  \caption{Per-component VRF on the Helmholtz problem ($d=100$):
    VRF per KL component (bars), sorted by decreasing prior eigenvalue
    $\lambda$ (orange), averaged over $20$ held-out observations. The
    dashed line marks $\text{VRF}=1$; the data-informed leading modes
    are hardest, while prior-dominated modes are reduced most.}
  \label{fig:helmholtz_component_vrf}
\end{figure}

\begin{figure}[t]
  \centering
  \includegraphics[width=\linewidth]{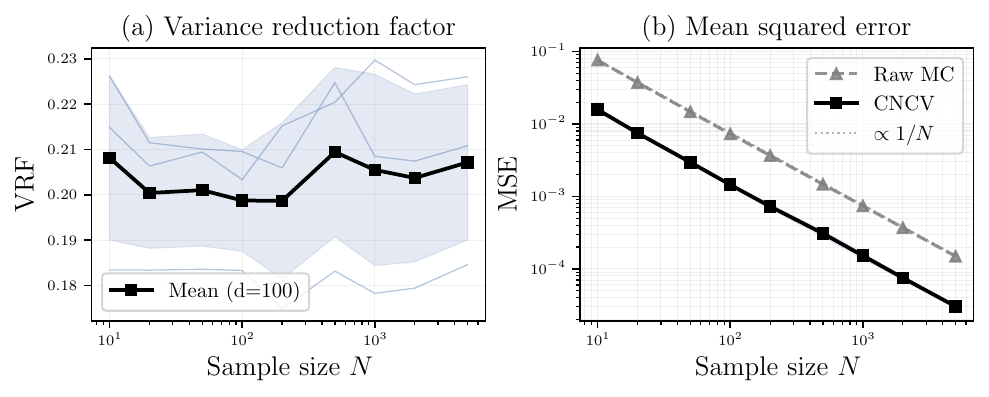}
  \caption{Sample efficiency on the Helmholtz problem.
    (a)~the mean VRF (over the $d=100$ components) is sample-size
    invariant; (b)~the CNCV estimator (solid) attains a
    constant-factor reduction in MSE over raw Monte Carlo (dashed),
    both following the $1/N$ rate.}
  \label{fig:helmholtz_sample_efficiency}
\end{figure}

Across $20$ held-out observations, CNCV attains a per-observation VRF
of $0.27 \pm 0.03$, below unity on all $20$---a ${\sim}\,3.7\times$
reduction in sample budget at fixed estimator MSE.
Figure~\ref{fig:helmholtz_posterior} shows the posterior summary for
one observation, Figure~\ref{fig:helmholtz_component_vrf} the
per-component VRF, and Figure~\ref{fig:helmholtz_sample_efficiency} the
sample efficiency. The empirical Stein-identity check also holds on
this second PDE: the dimension-averaged control-variate mean is
centered at zero with standard deviation ${\sim}\,10^{-3}$ across $250$
test observations, matching the four problems of the main text. CNCV
therefore reduces variance on a second, more oscillatory forward
operator with no change to the method.

\section{Experimental details and
  hyperparameters}\label{app:hyperparams}

We provide complete hyperparameter specifications for all experiments.
All models are trained with the Adam optimizer~\citep{KingmaBa2015}
unless otherwise noted (AdamW~\citep{LoshchilovHutter2019} for the
Darcy CNF). Learning rates
follow a cosine schedule from $\text{lr}_{\text{init}}$ to
$\text{lr}_{\text{final}}$.

\subsection{CNCV ensemble (Sections~\ref{sec:stylized}
  and~\ref{sec:darcy})}

\begin{table}[t]
\caption{CNCV ensemble hyperparameters.}
\label{tab:cncv_hparams}
\centering
\footnotesize
\setlength{\tabcolsep}{4pt}
\begin{tabular}{lcccc}
\toprule
& \textbf{Gauss.} & \textbf{Rosenb.} & \textbf{Nonlin.} & \textbf{Darcy} \\
\midrule
$d$ & 2--16 & 2 & 4 & 100 \\
$L$ (ensemble) & 16 & 16 & 16 & 16 \\
Tree depth & 1--3 & 2 & 2 & 3 \\
MLP layers & 5 & 3 & 3 & 5 \\
Hidden units & 128 & 64 & 64 & 128 \\
Batch size & 2048 & 2048 & 2048 & 4096 \\
Train samples & 65{,}536/131{,}072 & 65{,}536 & 65{,}536 & 120{,}000 \\
Epochs & 100 & 50 & 100 & 400 \\
$\text{lr}_{\text{init}}$ & $10^{-4}$ & $10^{-3}$ & $10^{-3}$ & $10^{-4}$ \\
$\text{lr}_{\text{final}}$ & $10^{-5}$ & $10^{-4}$ & $10^{-4}$ & $10^{-5}$ \\
Seed & 12 & 1 & 12 & 12 \\
\bottomrule
\end{tabular}
\end{table}

Table~\ref{tab:cncv_hparams} summarizes the CNCV hyperparameters.
For the Gaussian benchmark, the configuration is tuned per dimension:
the tree depth is $1, 2, 3, 3$ for $d = 2, 4, 8, 16$, and the training
set comprises $131{,}072$ samples for $d \le 8$ versus $65{,}536$ for
$d=16$ (the higher-data configuration did not train stably at
$d=16$). The Rosenbrock and nonlinear experiments share a single
architecture ($L\!=\!16$, depth~2, 64 hidden, 3-layer MLP), and the
Darcy problem requires a deeper MLP (5 layers) and lower learning rate
due to the higher dimensionality ($d\!=\!100$).

\subsection{Conditional normalizing flows
  (Sections~\ref{sec:stylized} and~\ref{sec:darcy})}

\begin{table}[t]
\caption{Conditional normalizing flow hyperparameters.}
\label{tab:cnf_hparams}
\centering
\footnotesize
\setlength{\tabcolsep}{4pt}
\begin{tabular}{lcccc}
\toprule
& \textbf{Gauss.} & \textbf{Rosenb.} & \textbf{Nonlin.} & \textbf{Darcy} \\
\midrule
Flow layers & 12 & 12 & 8 & 12 \\
Tree depth & 2 & 2 & 2 & 6 \\
MLP layers & 3 & 3 & 3 & 3 \\
Hidden units & 128 & 128 & 64 & 128 \\
Batch size & 8192 & 4096 & 8192 & 4096 \\
Train samples & 262{,}144 & 262{,}144 & 262{,}144 & 120{,}000 \\
Epochs & 1000 & 1000 & 1000 & 200 \\
$\text{lr}_{\text{init}}$ & $10^{-3}$ & $10^{-3}$ & $10^{-3}$ & $10^{-3}$ \\
$\text{lr}_{\text{final}}$ & $10^{-5}$ & $10^{-5}$ & $10^{-5}$ & $10^{-5}$ \\
Weight decay & 0 & 0 & 0 & $10^{-4}$ \\
Parameters & 0.6M & 0.2M & 0.1M & 17.2M \\
\bottomrule
\end{tabular}
\end{table}

Table~\ref{tab:cnf_hparams} shows the CNF hyperparameters. The
Gaussian and Rosenbrock CNFs use 12 flow layers with 128 hidden units,
while the nonlinear CNF uses 8 flow layers with 64 hidden units; all
are trained on 262k joint samples. The Darcy CNF is
substantially larger (12 flow layers, 128 hidden, 17.2M parameters) to
handle the 100-dimensional KL space with 33-dimensional observations.
Weight decay of $10^{-4}$ is applied only for the Darcy CNF, where it
prevents coupling scale explosion (Appendix~\ref{app:cnf_quality}).

\subsection{Diffusion model for the nonlinear experiment}

The denoising diffusion probabilistic model used for posterior
sampling in the nonlinear experiment
(Section~\ref{sec:stylized}) uses 1000 diffusion timesteps with
a linear noise schedule $\beta_t \in [10^{-4}, 0.02]$.
The denoising network is a 5-layer MLP with 512 hidden
units, conditioned on the observation $\mathbf{y}$ via
concatenation and on the timestep via sinusoidal position
embeddings. Training uses 65{,}536 joint samples, batch size
4096, and a cosine learning rate schedule from $10^{-3}$ to
$10^{-4}$ over 1000 epochs.

\subsection{Computational cost}

Training the CNCV ensemble takes approximately 5 minutes for the
stylized experiments ($d \leq 16$) and 50 minutes for the Darcy
problem ($d = 100$) on a single NVIDIA RTX 2000 Ada Generation GPU (16\,GB). The Darcy CNF
requires approximately 2 hours of training.
At inference time, evaluating $\mathbf{g}_{\text{ens}}$ for a
batch of $M$ samples requires $L = 16$ forward passes through
the coupling layer networks plus $M$ score evaluations.
For the Darcy problem with $M = 200$ samples, inference takes
${\sim}\,0.5$\,s, compared to ${\sim}\,100$\,s for 200 PDE solves.

\section{Comparison with classical Stein control
  variates}\label{app:gmm}

The control variates surveyed in Section~\ref{sec:related} are
non-amortized and, outside neural parameterizations, restricted to
fixed polynomial or kernel function classes. To examine how a low-order
polynomial basis behaves as the dimension grows and the posterior
becomes multimodal---the regime in which such a basis is least
expressive---we compare CNCV against degree-one and degree-two
polynomial Stein control variates~\citep{MiraSolgiImparato2013} and a
per-observation neural Stein control
variate~\citep{BedaqueOh2024, Oh2025}.

The target is an equally weighted two-component diagonal Gaussian
mixture, with modes separated along every coordinate and per-component
scale $0.3$, evaluated at $d \in \{10, 32, 64\}$ over three independent
repeats. CNCV uses the shared-hierarchical ensemble of
Section~\ref{sec:cncv}; the polynomial and per-observation neural
baselines are fit separately for each observation.

\begin{figure}[t]
  \centering
  \includegraphics[width=\linewidth]{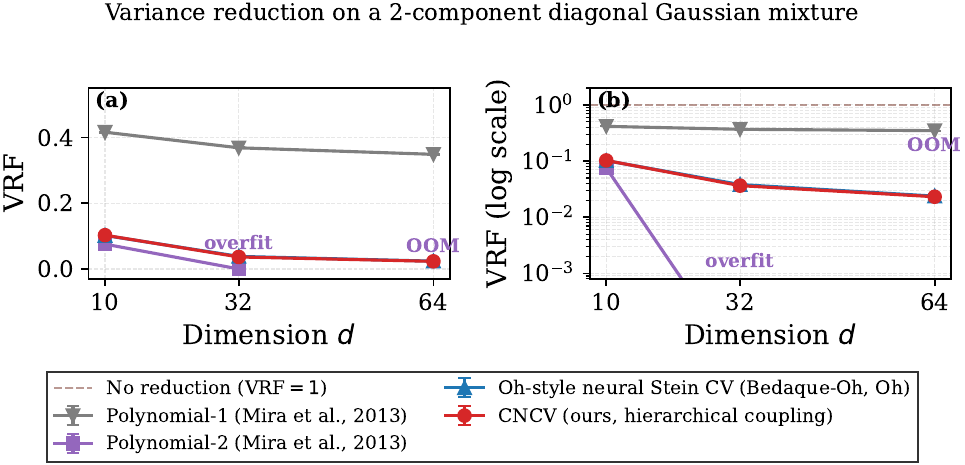}
  \caption{Variance reduction on a bimodal Gaussian mixture versus
    dimension~$d$. The degree-one polynomial control variate saturates
    near $0.4$ and does not improve with dimension. The degree-two
    value at $d=32$ (${\sim}\,2\times 10^{-5}$) is an overfit in-sample
    estimate, fit without a train/test split, and is unavailable at
    $d=64$ (out of memory). CNCV reaches $\approx 0.023$ at $d=64$,
    matching the per-observation neural variant while amortizing across
    observations.}
  \label{fig:gmm_scaling}
\end{figure}

Figure~\ref{fig:gmm_scaling} reports the VRF against dimension. The
degree-one polynomial control variate saturates near
$\text{VRF}\approx 0.4$ and does not improve with dimension, since a
linear function of the score cannot capture the bimodal posterior. The
degree-two variant is unreliable in this regime: at $d=32$ its
in-sample VRF collapses to ${\sim}\,2\times 10^{-5}$, but this value
reflects overfitting of its $\mathcal{O}(d^2)$ coefficients rather than
genuine variance reduction, as it is fit without a train/test split; at
$d=64$ the quadratic feature expansion exhausts memory and no estimate
is available. By contrast, both the per-observation neural Stein
control variate and CNCV reduce variance increasingly with dimension,
reaching $\text{VRF}\approx 0.023$ at $d=64$ with negligible variation
across repeats. CNCV matches the per-observation neural variant
(neural $0.024$ versus CNCV $0.023$ at $d=64$) while requiring a single training
run that generalizes across observations, whereas the neural baseline
is refit for every observation.

\section{Exact divergence versus stochastic trace
  estimation}\label{app:hutchinson}

The hierarchical coupling architecture of Section~\ref{sec:cncv}
returns the divergence $\nabla_\mathbf{x} \cdot \boldsymbol{\phi}$ exactly, as the
sum of the Jacobian diagonal computed in a single forward pass. The
generic alternative for a black-box $\boldsymbol\phi$ is the Hutchinson
estimator~\citep{Hutchinson1990}, which approximates the trace with $k$
random probe vectors. We quantify the accuracy and cost of this
alternative against the exact diagonal.

\begin{figure*}[tb]
  \centering
  \includegraphics[width=0.92\textwidth]{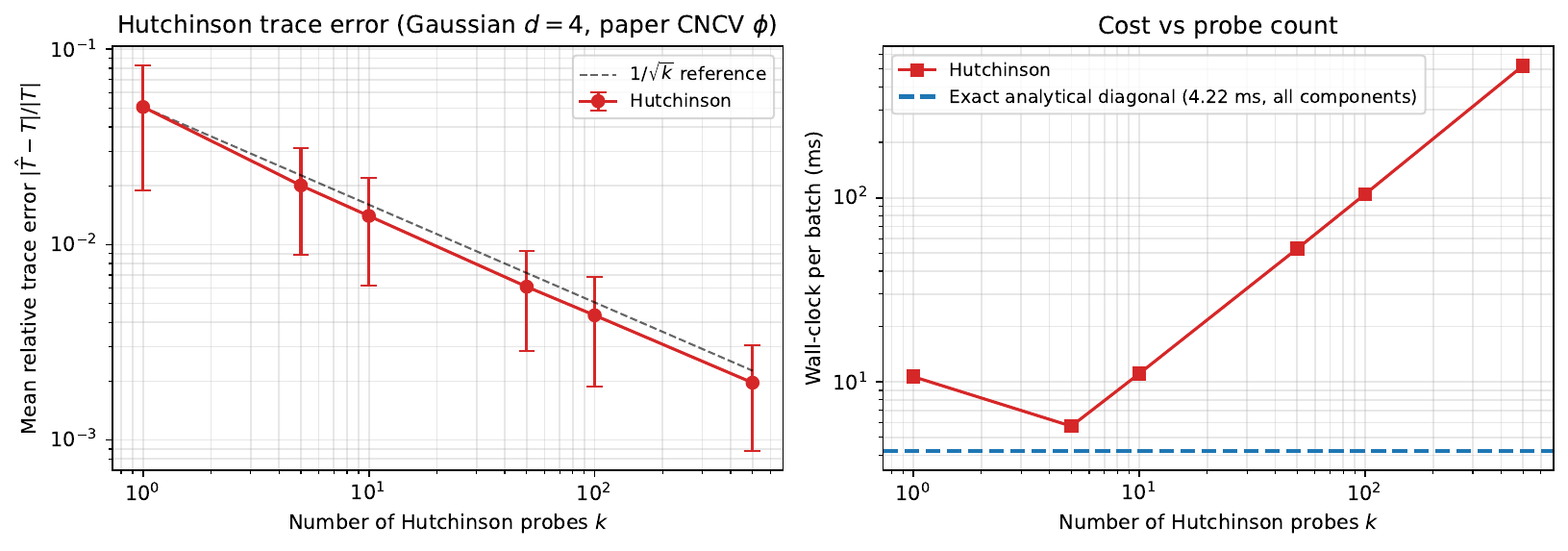}
  \caption{Relative trace error (left axis) and wall-clock time
    (right axis) of the Hutchinson estimator versus probe count~$k$ on
    the Gaussian problem ($d=4$), against the exact Jacobian diagonal.}
  \label{fig:hutchinson_d4}
\end{figure*}

\begin{figure*}[tb]
  \centering
  \includegraphics[width=0.92\textwidth]{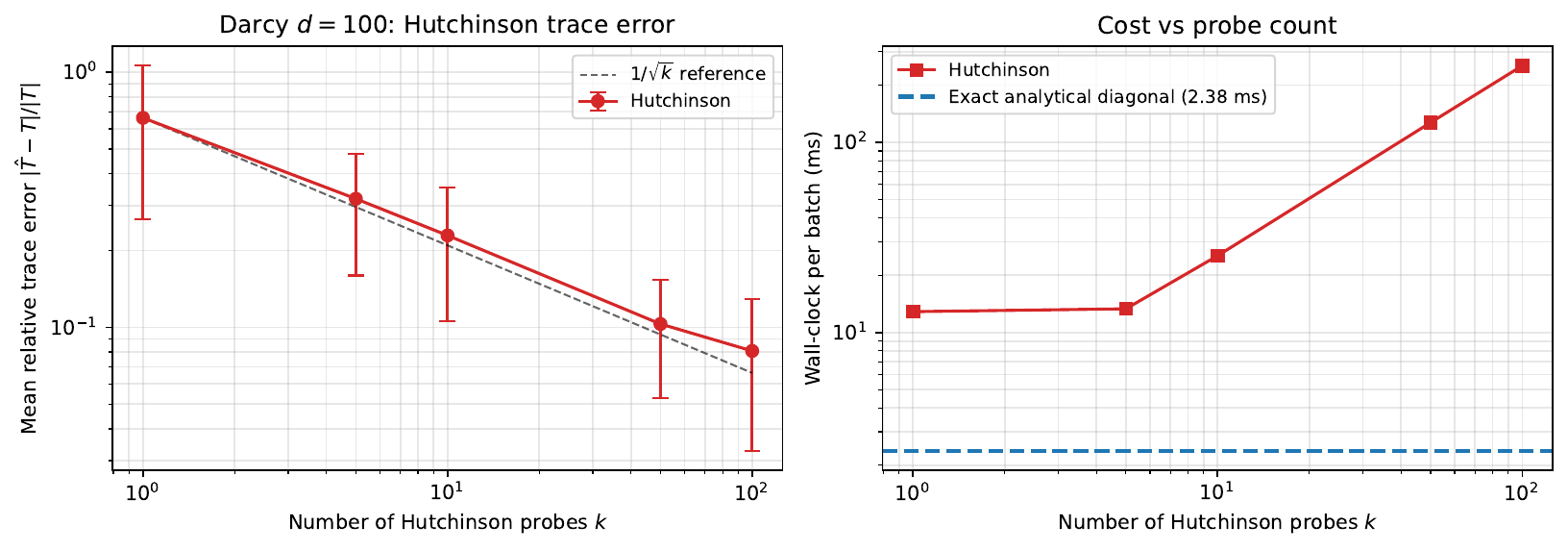}
  \caption{The same comparison on the Darcy problem ($d=100$); the
    Hutchinson trace error and its cost relative to the exact diagonal
    both grow with dimension.}
  \label{fig:hutchinson_darcy}
\end{figure*}

Figures~\ref{fig:hutchinson_d4} and~\ref{fig:hutchinson_darcy} report
the relative trace error and wall-clock time as a function of the probe
count~$k$. On the Gaussian problem ($d=4$), the exact diagonal is
computed in ${\sim}\,4$\,ms; matching its accuracy with the Hutchinson
estimator requires $k\approx 100$ probes at ${\sim}\,105$\,ms---roughly
$25\times$ the cost---and still leaves a relative error of
${\sim}\,0.4\%$. The gap widens sharply with dimension. On the Darcy
problem ($d=100$), a single probe incurs a 66\% relative error, and
$k=100$ probes still leave an 8\% error at ${\sim}\,106\times$ the cost
of the exact diagonal. A second $d=100$ PDE-constrained problem---full
waveform inversion governed by the Helmholtz equation, with the same KL
parameterization and a learned CNF score---shows the same pattern:
$k=100$ probes leave an 11\% error at ${\sim}\,109\times$ the cost. This widening is expected,
since the variance of the Hutchinson estimator scales with the squared
Frobenius norm of the Jacobian, which grows with dimension. The exact
diagonal is therefore not merely a convenience but a prerequisite for
tractable, low-variance divergence estimation at these scales.

\section{Non-Gaussian observation noise}\label{app:studentt}

All experiments in Section~\ref{sec:experiments} use Gaussian
observation noise. Because the control variate is built from the
posterior score rather than any Gaussian assumption
(Section~\ref{sec:cncv}), it applies unchanged to non-Gaussian
likelihoods. We demonstrate this on a heavy-tailed Student-$t$
likelihood.

We take a standard Gaussian prior in $d=4$ and a component-independent
Student-$t$ likelihood with $\nu=5$ degrees of freedom and scale
$0.3$, for which the posterior score is available in closed form. For
each of $20$ held-out observations we draw posterior samples by
Langevin dynamics and evaluate the per-observation VRF, comparing CNCV
against a per-observation degree-one polynomial Stein control variate
fit by least squares.

\begin{figure}[t]
  \centering
  \includegraphics[width=\linewidth]{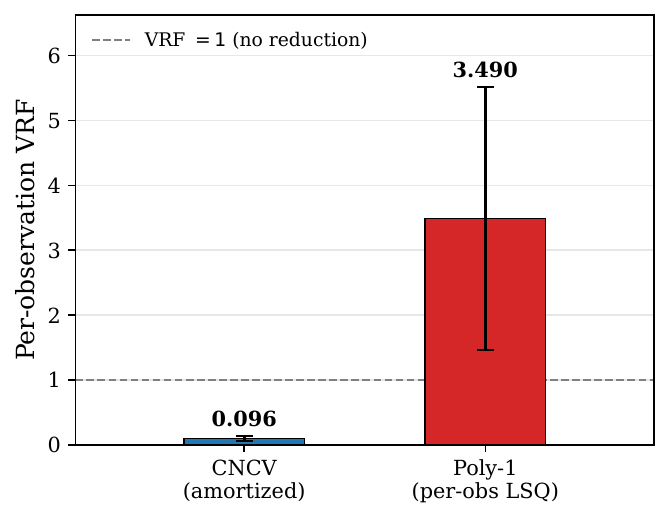}
  \caption{Per-observation VRF under a Student-$t(\nu=5)$ likelihood
    ($d=4$, 20 held-out observations). CNCV reduces variance on every
    observation (mean $0.10$), whereas the per-observation linear
    polynomial Stein control variate increases variance on 95\% of
    them (mean $3.5$). The optimal control variate is nonlinear in the
    saturating Student-$t$ score.}
  \label{fig:studentt}
\end{figure}

Figure~\ref{fig:studentt} shows the per-observation results. CNCV
reduces variance on every observation (mean VRF $0.10 \pm 0.04$),
whereas the degree-one polynomial control variate increases variance on
all but one (mean $3.5$): the Student-$t$ score saturates for large
residuals, so the optimal control variate is strongly
nonlinear---beyond the reach of a linear basis but captured here by the
coupling-layer ensemble, the same mechanism that favors CNCV on the
non-Gaussian posteriors of Section~\ref{sec:stylized}.

\section{Robustness to score approximation error}\label{app:score_pert}

Section~\ref{sec:stylized} observed that a learned CNF score yields
variance reduction comparable to the analytical score. To probe
this robustness directly and in isolation, we perturb the analytical
Gaussian score ($d=4$) with isotropic Gaussian noise of increasing
magnitude, hold the trained CNCV ensemble fixed, and measure the
resulting VRF over $100$ observations.

\begin{figure}[tb]
  \centering
  \includegraphics[width=\linewidth]{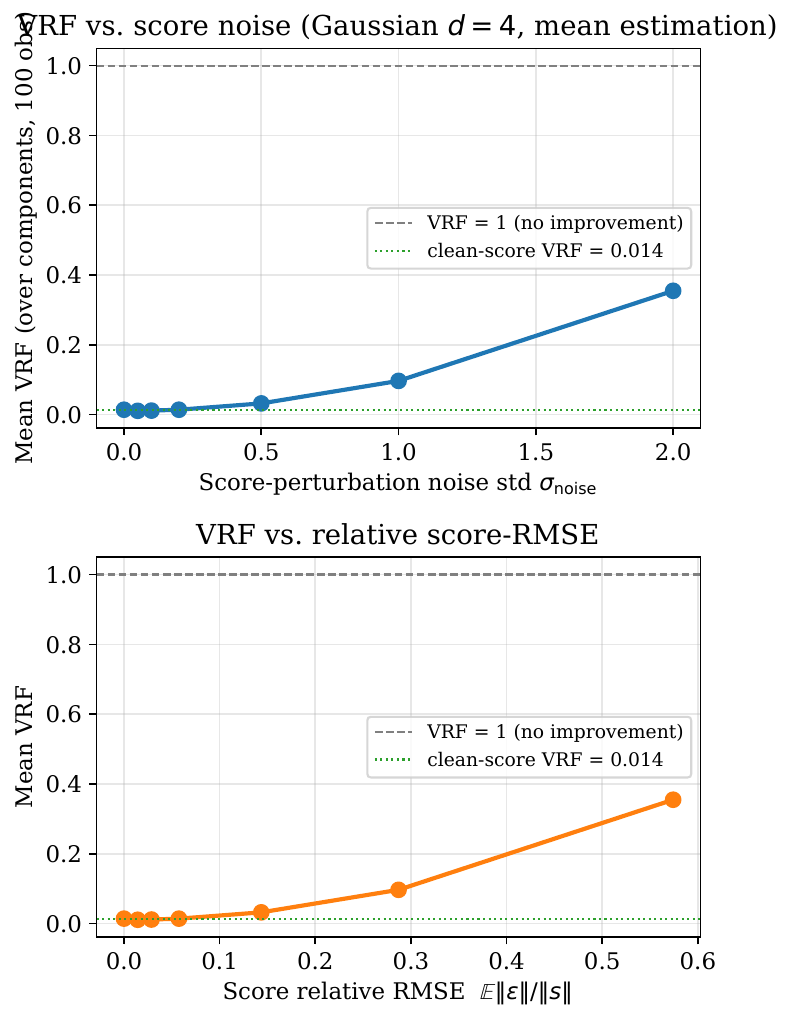}
  \caption{VRF versus relative score error on the Gaussian problem
    ($d=4$), perturbing the analytical score with isotropic Gaussian
    noise while holding the CNCV ensemble fixed. The VRF degrades
    gracefully: a 14\% error yields $0.032$, and a 57\% error still
    yields $0.35$.}
  \label{fig:score_pert}
\end{figure}

Figure~\ref{fig:score_pert} shows that the VRF degrades gracefully and
monotonically with the relative score error. A 14\% score error
raises the VRF only from $0.014$ to $0.032$, and even a 57\% error
leaves it at $0.35$---still a substantial reduction, consistent with
the learned-score result of Section~\ref{sec:stylized}. This
isotropic perturbation does not reproduce the structured error of a
learned score, but it isolates the control variate's sensitivity to
score inaccuracy.

\end{document}